%% file: 0.sample-sigconf.tex
\documentclass[sigconf, nonacm]{acmart}

\usepackage{enumitem}
\usepackage{multirow}

\usepackage{booktabs}
\usepackage{multirow}
\usepackage{makecell}
\usepackage{array}
\usepackage{caption}
\usepackage{graphicx}
\usepackage{colortbl}
\usepackage{subcaption}
\usepackage{siunitx} 
\usepackage{tcolorbox}
\tcbuselibrary{breakable}

\usepackage{algorithm}
\usepackage{algpseudocode}

\AtBeginDocument{%
  }

\setcopyright{rightsretained} 
\copyrightyear{2026}
\acmYear{2026}
\acmDOI{XXXXXXX.XXXXXXX}
\acmConference[KDD '2026]{the 32st ACM SIGKDD Conf. on Knowledge Discovery and Data Mining}{August 9-13, 2026}{Jeju, Korea}
\acmISBN{978-1-4503-XXXX-X/2026/06}

\makeatletter
\patchcmd{\maketitle}{\@copyrightpermission}{
   \begin{minipage}{0.3\columnwidth}
     \href{https://creativecommons.org/licenses/by/4.0/}{\includegraphics[width=0.90\textwidth]{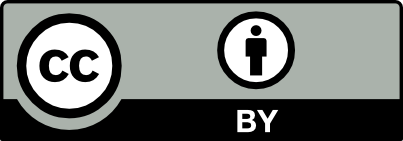}}
   \end{minipage}\hfill
   \begin{minipage}{0.7\columnwidth}
     \href{https://creativecommons.org/licenses/by/4.0/}{This work is licensed under a Creative Commons Attribution International 4.0 License.}
   \end{minipage}
   \vspace{5pt}
}{}{}
\makeatother

\begin{document}

\title{Dialogue Model Optimization via Agent Game and Adaptive Tree-based GRPO}

\author{Kun Peng$^\dagger$\textsuperscript{a}, Conghui Tan$^\ddagger$, Yu Liu$^\dagger$, Guohua Tang$^\ddagger$, Zhongqian Sun$^\ddagger$, Wei Yang$^\ddagger$, Zining Zhu$^\ddagger$, \\Lei Jiang$^\dagger$\textsuperscript{b}, Yanbing Liu$^\dagger$, Hao Peng$^\dagger$}

\affiliation{%
 \institution{$^\dagger$Institute of Information Engineering, Chinese Academy of Sciences
 $^\ddagger$Tencent
 }\country{}
}
\email{pengkun@iie.ac.cn}

\thanks{
\par\textsuperscript{a} Work done during internship at Tencent
\par\textsuperscript{b} Corresponding author}

\renewcommand{\shortauthors}{Kun et al.}

\begin{abstract}

Open-ended dialogue agents aim to deliver engaging, personalized interactions by adapting to users' traits, but existing methods face critical limitations: over-reliance on pre-collected user data, and short-horizon biases in reinforcement learning (RL) that neglect long-term dialogue value. To address these, we propose a novel long-horizon RL framework integrating online personalization with Adaptive Tree-based Group Relative Policy Optimization (AT-GRPO). Adopting a two-agent game paradigm, a user agent constructs dynamic environments via style mimicry (learning user-specific conversational traits) and active termination (predicting turn-level termination probabilities as immediate rewards), forming an iterative cycle that drives the dialogue agent to deepen interest exploration. AT-GRPO reinterprets dialogue trajectories as trees and introduces adaptive observation ranges. Unlike full tree expansion that incurs exponential overhead, it limits each node to aggregate rewards from a stage-aware range: larger ranges support early-stage topic exploration, while smaller ranges facilitate late-stage dialogue maintenance. This design reduces rollout budgets from exponential to polynomial in the dialogue length, while preserving long-term reward capture. Extensive experiments show our framework's superior performance, sample efficiency, and robustness.

\end{abstract}

\ccsdesc[500]{Computing methodologies~Discourse, dialogue and pragmatics}
\ccsdesc[500]{Computing methodologies~Reinforcement learning}

\keywords{Dialogue Model, Reinforcement Learning}

\maketitle

\section{Introduction}
Open-ended dialogue agents built on large language models (LLMs) have grown increasingly indispensable in human-computer interaction, finding broad applications in social robotics, game non-player characters (NPCs), virtual assistants, and educational mentors \cite{graesser2004autotutor, fitzpatrick2017delivering, adewumi2022state, zhou2020design, zhou-etal-2024-characterglm}. Beyond generating grammatically sound responses, the primary objective of such agents is to flexibly adapt to users’ distinct personalities, interest inclinations, and conversational patterns—delivering a personalized interaction experience that strengthens user involvement and loyalty \cite{poddar2024personalizing, wu2025rlpf, chen2025pal}. Despite remarkable advancements in LLMs that have elevated dialogue fluency and contextual comprehension \cite{tseng-etal-2024-two}, their capacity to achieve effective personalization in real-world scenarios remains limited, particularly when no prior user data is accessible.

Personalization is essential rather than optional for human-oriented conversational systems.
However, the majority of existing personalization approaches for LLMs depend heavily on large volumes of pre-collected user histories, profiles, or latent feature representations \cite{tan-etal-2024-personalized, tan-etal-2024-democratizing, zhu-etal-2024-inters, salemi-etal-2024-lamp, salemi2024optimization, liu-etal-2025-llms}, restricting their applicability in on-demand scenarios such as anonymous interactions, temporary virtual assistants, or privacy-protected domains. These scenarios call for online personalization capabilities---where the agent acquires insights into the user dynamically during the conversation, reducing ambiguity about user characteristics as the interaction progresses without relying on offline data stores \cite{wan2025enhancing}.

\begin{figure}[t]
    \centering
    \includegraphics[width=0.85\linewidth]{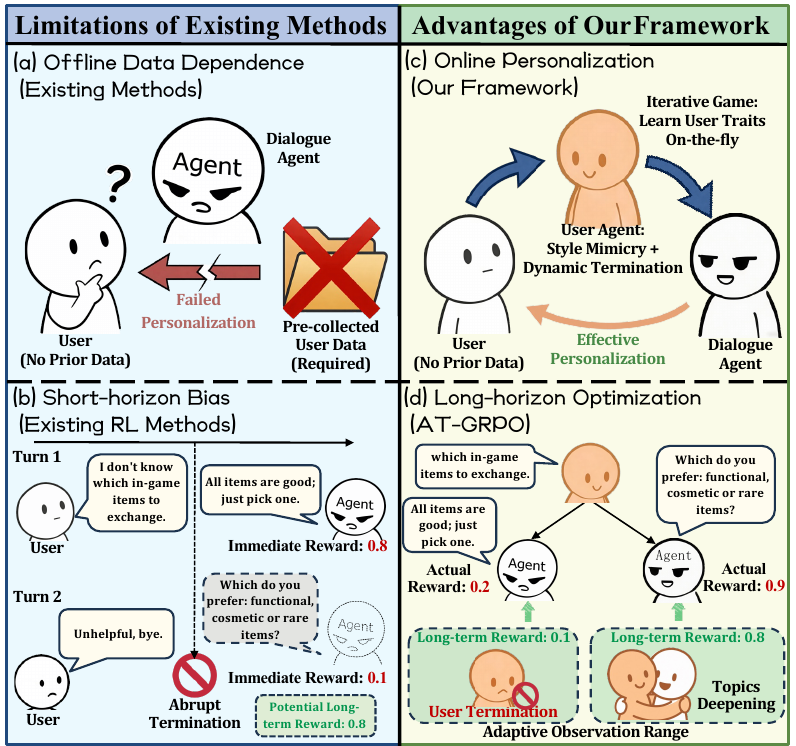}
    \caption{Limitations of existing methods vs. our framework.}
    \label{fig:1}
\end{figure}

Reinforcement Learning (RL) has emerged as a viable approach for training personalized dialogue agents, enabling them to refine their strategies through real-time feedback from interactions \cite{ouyang2022training}. That said, existing RL techniques for dialogue generation, including PPO \cite{schulman2017proximal} and GRPO \cite{shao2024deepseekmath}, are plagued by short-horizon optimization biases—they prioritize immediate rewards while overlooking the long-term value of conversations. As shown in Figure \ref{fig:1}, this myopia gives rise to two key problems: first, the "immediate reward trap," where high instant feedback leads to abrupt dialogue termination due to inadequate exploration of user interests; second, the "undervaluation of exploratory actions," where utterances with low immediate rewards but the potential to uncover deep user preferences are abandoned, squandering chances for sustained high-quality interactions. Furthermore, chain-based rollout strategies in traditional RL methods result in redundant token usage and sparse supervision signals, as trajectory-level outcome rewards fail to isolate the contribution of individual dialogue turns \cite{hou2025treerl, ji2025tree, yang2025treerpo}. 

Another major hurdle in RL-driven dialogue personalization is reward hacking---agents may employ opportunistic tactics (e.g., meaningless repetitive queries, forced conversation retention, or directive responses) to prolong dialogue duration without enhancing content quality, which distorts reward signals and undermines model optimization \cite{10.5555/3600270.3600957}. To make matters worse, existing online personalization methods lack dynamic interaction environments that can simulate real user behaviors, as static dataset-based imitation learning is unable to adapt to diverse user types or evolving conversational contexts \cite{jiang2025know}. This research gap underscores the need for an integrated framework that addresses long-horizon optimization, dynamic online personalization, and robust reward design simultaneously.

To overcome these limitations, we propose a novel long-horizon RL framework based on agent game theory, integrating online personalization mechanisms with Adaptive Tree-based Group Relative Policy Optimization (AT-GRPO) to boost the adaptive capabilities of dialogue agents. Our framework utilizes a two-agent game between a dialogue agent and a user agent: the latter constructs a dynamic interaction environment, while the former optimizes its strategies to explore user traits and sustain high-quality long-term dialogues.

The user agent is equipped with two core mechanisms to enable realistic dynamic interactions. First, the style mimicry mechanism learns user-specific conversational styles, preferences, and response logic by training on real user dialogue corpora through supervised fine-tuning (SFT), thereby generating responses that align with target user characteristics and supporting adaptive simulation of diverse user profiles. 
Second, within the active dialogue termination mechanism, the user agent predicts a turn-level termination probability based on conversation history; this probability serves both as an explicit feature for the next dialogue turn and as an immediate reward signal for the conversational agent. This mechanism can accurately identify natural termination cues (such as topic exhaustion or declining user interest), forming a game-theoretic cycle: the user agent gradually raises its termination threshold to demand higher interaction quality, while the conversational agent optimizes its strategy by delving deeper into user interest exploration, thereby reducing the termination probability.

To address short-horizon biases and inefficient rollouts, our AT-GRPO reinterprets dialogue trajectories as tree structures—each node denotes a dialogue turn, and child nodes represent sampled candidate responses \cite{hou2025treerl, ji2025tree, yang2025treerpo}. The original TreeRPO \cite{yang2025treerpo} addresses short-horizon biases by adopting bottom-up weighted reward aggregation, enabling each node to capture cumulative rewards from all subsequent paths. However, this comprehensive aggregation comes at a steep cost: expanding all nodes in the tree results in exponential growth of computational overhead, which is impractical for real-world dialogue systems with limited computational resources.
To resolve this inefficiency while retaining the long-horizon reward capture capability, AT-GRPO introduces an adaptive expansion mechanism. Instead of requiring each node to access all subsequent child nodes (full tree expansion), we limit each node to an adaptive observation range—only nodes within this range are included in the bottom-up weighted reward aggregation. This design reduces the original rollout budget from exponential growth to polynomial, drastically lowering computational costs without sacrificing key performance. The observation range is adaptively adjusted based on the dialogue stage: in the early stages, nodes need to explore high-value topics to lay the foundation for sustained interaction, so a larger observation range is adopted to capture more potential long-term rewards; in the later stages, the focus shifts to maintaining the ongoing dialogue and deepening existing topics, where a smaller observation range suffices as complete supervision signals are no longer necessary. This stage-aware adaptive mechanism ensures that computational resources are allocated efficiently while preserving the ability to balance immediate feedback and long-term gains.


We validate our framework on multiple benchmark datasets, including LCCC \cite{wang2020large}, DailyDialog \cite{li2017dailydialog}, and a custom game NPC dataset, evaluating performance across dimensions such as interaction length, dialogue coherence, and user satisfaction. Our experimental results demonstrate that the proposed framework outperforms state-of-the-art baselines in online personalization and long-horizon interaction, achieving impressive performance with only 100 training steps.

In summary, our contributions are threefold: 

1. We propose an agent game framework for online personalization, where the user agent integrates style mimicry and active termination capabilities to build dynamic interaction environments, enabling the conversational agent to learn user traits through iterative game-based interactions without relying on prior data.

2. We develop AT-GRPO for dialogue optimization, leveraging tree-structured rollouts and weighted reward aggregation to eliminate short-horizon biases, and enhance sample efficiency.


3. We establish a comprehensive evaluation protocol across diverse domains, providing empirical proof that our framework outperforms baseline methods in personalization, long-horizon interaction, and robust domain adaptation. 


\input{1.relatedwork}
\input{2.method}
\input{3.experement}

\section{Conclusion}
In this paper, we propose a novel long-horizon RL framework for dialogue model optimization, integrating two-agent game-based online personalization with Adaptive Tree-based Group Relative Policy Optimization (AT-GRPO). The user agent constructs dynamic interaction environments via style mimicry and active termination, enabling the dialogue agent to learn user traits on-the-fly without relying on pre-collected data. AT-GRPO addresses short-horizon biases by modeling dialogue trajectories as trees with adaptive observation ranges, reducing computational overhead from exponential to polynomial growth while preserving long-term reward capture. Extensive experiments on three datasets demonstrate that our framework outperforms state-of-the-art baselines in personalization, long-horizon interaction, and generalization, with superior sample efficiency and computational feasibility. Future work will explore extending the framework to multi-modal dialogue scenarios and incorporating more fine-grained user trait modeling.

\newpage
\bibliographystyle{ACM-Reference-Format}
\bibliography{bibfile}

\appendix
\input{4.appendix}

\end{document}

%% file: 1.relatedwork.tex
\section{Related Work}\label{sec:relatedwork}

\begin{figure*}[ht]
    \centering
    \includegraphics[width=0.95\linewidth]{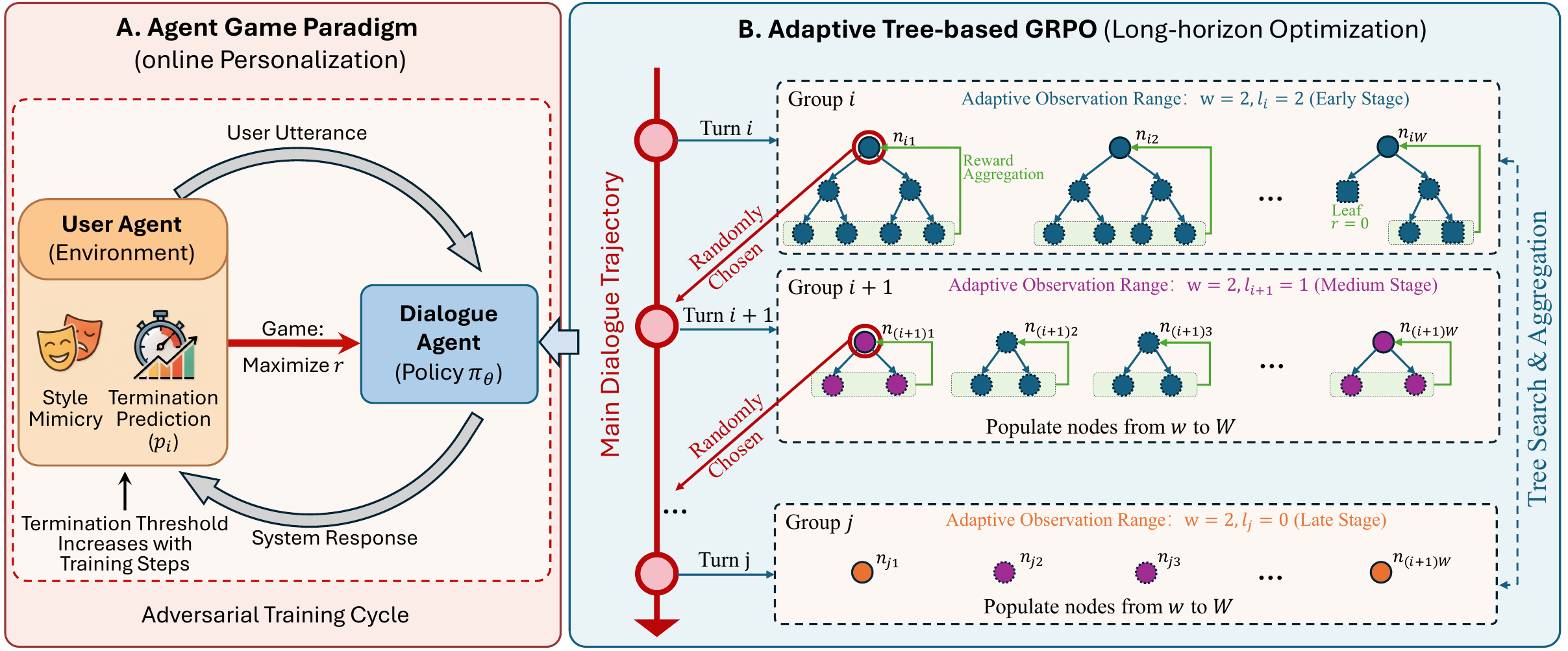}
    \caption{The overall framework operates through an iterative closed loop: the user agent provides real-time feedback (e.g., termination probability as reward signals) based on the dialogue agent’s responses, and the dialogue agent is trained via AT-GRPO to balance immediate interaction quality and long-term conversational value.}
    \label{fig:overall}
\end{figure*}

\subsection{Personalized Dialogue LLM}
Personalized LLMs generate high-quality content by leveraging user-specific data, and existing approaches are primarily categorized into two paradigms:
Fine-tuning methods \cite{tan-etal-2024-personalized, tan-etal-2024-democratizing, zhu-etal-2024-inters} optimize partial model parameters for personalization. PER-PCS \cite{tan-etal-2024-personalized} uses LoRA to fine-tune user-specific Llama models, while OPPU \citep{tan-etal-2024-democratizing} reduced costs via user clustering. However, these methods rely on user-specific data and face challenges in dynamic preference adaptation.
Retrieval-based methods \cite{salemi-etal-2024-lamp, salemi2024optimization, liu-etal-2025-llms} inject personalization through in-context demonstrations. LaMP \cite{salemi-etal-2024-lamp} retrieved relevant user histories, and ROPG-RL \cite{salemi2024optimization} optimized retrievers via RL. Yet, they prioritize task relevance over holistic user patterns, limiting long-term preference capture.
Distinct from these offline data-dependent methods, CURIO \cite{wan2025enhancing} pioneers online personalization via curiosity-driven rewards, but it still relies on predefined discrete user types. This constraint renders the model ill-suited for open-domain conversations, where user preferences are often continuous, ambiguous, or dynamically evolving (e.g., users may temporarily mention new interests during interactions). Such predefined categorization fails to cover real-world preference complexity, limiting generalization.

\subsection{Tree-Search Enhanced RL for LLM}
Tree search has emerged as a promising auxiliary paradigm for reinforcing reasoning capabilities in LLMs, with existing efforts primarily focusing on three directions: inference-time enhancement, offline data synthesis, and limited attempts at online RL integration.
Inference-time tree search methods \cite{snell2024scaling, yao2023tree, long2023large} leverage structured exploration to improve response quality during decoding,
but these methods rely on external reward signals or heuristic evaluation, lacking tight integration with RL training and failing to transfer exploration gains to model parameter updates.
Offline training with tree search \cite{xie2024monte, chen2024alphamath, zhang2024rest} focuses on generating high-quality training data for supervised fine-tuning (SFT) or direct preference optimization (DPO) \cite{rafailov2023direct}.
However, they suffer from distribution mismatch and reward hacking \cite{guo2025deepseek} when applied to offline RL, as the static process reward models derived from tree search fail to adapt to the dynamic policy shifts during training.

Limited works have explored integrating tree search into online RL training \cite{feng2023alphazero, kazemnejad2024vineppo}. 
Unlike AlphaZero \cite{silver2017mastering} which tightly couples tree search with RL in game AI, LLM RL has not fully exploited tree search's ability to provide adaptive process rewards and efficient exploration. This disconnect limits the model's capacity to learn from structured reasoning paths and transfer exploration gains to long-term performance improvements.

%% file: 2.method.tex
\section{Proposed Methodology}\label{sec:method}


\subsection{User Agent}
The overall framework is shown in Figure~\ref{fig:overall}. The user agent serves as a dynamic, adaptive interaction environment that simulates real user behaviors, including conversational style mimicry and active dialogue termination. It is trained via SFT on real user dialogue corpora and provides explicit reward signals for the dialogue agent during RL.

\subsubsection{Style Mimicry and Termination Preference Learning}
High-quality SFT data is crucial for user agent training.
We thus collect and select a subset of multiple conversational data entries from the same real user as prompt demonstrations, with the remaining conversational data used as training samples. 
This setup enables the user agent to learn the mapping from the user’s traits to their conversational style from the historical multi-turn conversational data.
To learn the termination preference, the user agent is trained with a constrained output format: each response must start with either the token \([\textit{Continue}]\) or \([\textit{Terminate}]\), which explicitly indicates the user's willingness to continue the dialogue.


\subsubsection{Explicit Probability Propagation and Dynamic Threshold Adjustment}
In the agent game, the user agent's parameters are frozen to maintain a stable interaction environment. 
We define the interaction trajectory as $H=[D_0, U_0, D_1, U_1, ...,D_N, U_N]$, where $D_i \in \mathcal{D}$ (dialogue agent's utterances) and $U_i \in \mathcal{U}$ (user agent's utterances).
The interaction terminates when the user agent outputs [\textit{Terminate}] ($U_N=[\textit{Terminate}]$).
At each interaction turn, the user agent can compute the termination probability $p_i$ in real time without relying on any additional parametric modules:
\begin{equation}
    p_i = \min\left( \alpha \times 2 P([\textit{Terminate}] | H_{<i}, D_i), 1 \right),
\end{equation}
where \( H_{<i} = [D_0, U_0, ..., D_{i-1}, U_{i-1}] \) is the historical context up to turn \( i-1 \), \( P(\cdot) \) denotes the probability output by the user agent. $\alpha$ controls the strictness of the user agent's termination preference.
When \( P([\textit{Terminate}]|H_{\leq i},D_i) \geq 0.5 \), \( p_i = 1 \), indicating the user will terminate the dialogue.
Termination probability $p_i$ serves two key functions:

\textbf{Explicit Probability Propagation}: 
To enable the user agent to track historical termination tendencies, we propagate the termination probability as an explicit feature in the dialogue context. Specifically, the user's \( (i+1) \)-th response \( U_{i+1} \) is generated based on the augmented context that includes all previous termination probabilities:
\begin{equation}
U_{i+1} = P_{\text{user}} \left( U \mid D_0 \oplus p_{init}, U_0, D_1 \oplus p_0, ..., U_i, D_{i+1}\oplus p_i \right),
\end{equation}
where \( p_{init} = 0 \) (initial termination probability) and \( D_{i+1} \oplus p_i \) denotes appending \( p_i \) to the dialogue agent's \( i+1 \)-th utterance. This augmentation allows the user agent to learn context-aware termination decisions.

\textbf{Dynamic Threshold Adjustment}: 
The immediate reward signal for the dialogue agent is defined as \( r_i = 1 - p_i \) (higher rewards indicate stronger user willingness to continue).
To form an adversarial training cycle, we dynamically adjust \( \alpha \) linearly with RL progress:
\begin{equation}
\alpha = 1 + \lambda \cdot \left\lfloor \frac{\textit{steps}}{10} \right\rfloor,
\end{equation}
where \( \textit{steps} \) is the current RL training step, and \( \lambda > 0 \) is a hyperparameter regulating the rate of strictness increase. As training proceeds, the user agent becomes less tolerant of low-quality interactions, driving the dialogue agent to continuously improve its strategy.

\subsection{Dialogue Agent and Adaptive Tree-based GRPO (AT-GRPO)}
The dialogue agent is first initialized via SFT on high-quality dialogue datasets to acquire basic conversational competence (e.g., fluency, contextual coherence). 
Beyond generating high-quality responses, its core goal during RL is to proactively explore user interests and optimize its policy to minimize the user agent's termination probability \( p_i \) (i.e., maximize immediate rewards \( r_i \)) while sustaining long-term high-quality interactions.



Vanilla GRPO uses chain-based rollouts, where for each turn in the dialogue, it samples a group of size $W$, then selects one node from it as the main trajectory to continue sampling. The overall complexity is $\mathcal{O}(WL)$, where $L$ is the dialogue exchange length.
But this approach fails to capture long-term conversational value. To address this, our AT-GRPO model the dialogue trajectory as a tree structure $T$:




\begin{enumerate}[leftmargin=4pt, itemindent=*, label=$\bullet$]
    \item Each node \( n_{ij} \) represents the \( j \)-th sampled interaction at the \( i \)-th dialogue turn (depth \( i \)), corresponding to the triple \( (D_{ij}, U_{ij}, r_{ij}) \).
    \item The number of child nodes per parent node is \( W \) (GRPO's group sampling size), representing \( W \) candidate responses generated by the dialogue agent for the current context.
    \item A node \( n_{ij} \) is a leaf node if and only if $r_{ij}=0$ (i.e., $U_{ij}=[\textit{Terminate}]$) or the maximum depth $L$ is reached (maximum allowed dialogue exchange length).
\end{enumerate}

To let early nodes sense long-term value, one naive idea \cite{yang2025treerpo} is to first expand the entire tree, then backpropagate rewards from the leaves upward.
However, this approach is plagued by two key drawbacks.
\textbf{First}, because the number of nodes grows by a factor of $W$ at each depth, the RL process naturally focuses more on later dialogue stages. This biases the learning toward late-phase decisions and contradicts our goal of encouraging exploration in early phases.
\textbf{More critically}, full expansion of this tree requires \( \sum_{t=1}^L W^t \approx W^L \) interactions, leading to exponential computational overhead \( \mathcal{O}(W^L) \) that is infeasible for large \( W \) and \( L \).

To address these issues, our proposed AT-GRPO employs a layered exploration strategy: as shown on the right of Figure \ref{fig:overall}, at each layer, it first explores all subsequent nodes equally within an adaptive observation range, then randomly selects one non-leaf node to extend the main trajectory to the next layer. This ensures a uniform group size at every depth.

\subsubsection{Adaptive Observation Range}
We introduce an adaptive expansion with the insight that dialogue strategies vary with turn depth:

\textbf{Early turns (deep depth)}: The agent needs to explore high-value topics, requiring a larger observation range to capture potential long-term rewards;

\textbf{Late turns (shallow depth)}: The agent focuses on maintaining and deepening existing topics, prioritizing immediate rewards over distant future gains.

For a non-leaf node \( n_{ij} \) at depth \( i \), we define its observation length \( l_i \) (number of subsequent layers to expand) as:
\begin{equation}
l_i = \text{round}\left( \gamma \ln\left( L - i + 1 \right) \right), 
\end{equation}
where \( \gamma > 0 \) is a hyperparameter scaling the observation length. \( l_i \) decreases as \( i \) increases, adapting to the dialogue's strategic needs.
The expansion budget for the observation subtree rooted at node \( n_{ij} \), with width $w$ and depth $l_i$, is given by \( \sum_{t=1}^{l_i} w^t \approx w^{l_i} \).
The reward for node \( n_{ij} \) is a weighted combination of its immediate reward and the average reward of all leaf nodes within its observation range:
\begin{equation}
r'_{ij} = \omega r_{ij} + (1-\omega)\frac{1}{w^{l_i}} \sum_{k \in \mathcal{L}_{ij}^{(l_i)}} r_k
\end{equation}
where \( \mathcal{L}_{ij}^{(l_i)} \) denotes the set of all leaf nodes obtained from \( n_{ij} \) subtree, with \( |\mathcal{L}_{ij}^{(l_i)}| = w^{l_i} \), $\omega$ is a weight to trade off between immediate and long-term rewards.

\subsubsection{Single-trajectory Tree Expansion}
After aggregating rewards to form a GRPO group, we randomly select a non‑leaf node as the context for the next dialogue turn. 
Based on its subtree, we populate its children nodes from $w$ to $W$ to form a new GRPO group.
We then repeat the steps of adaptive expansion and reward aggregation until all nodes become leaf nodes or the maximum depth is reached.

\subsubsection{Computational Complexity}

The budget for expanding the entire GRPO tree is:
\begin{equation}
\label{eq-xx}
\sum_{i=1}^{L}{W\sum_{t=1}^{l_i}w^t} \leq b \sqrt{w}W \sum_{k=1}^{L} k^{\gamma \ln w}
\leq b \sqrt{w}W L^{1 + \gamma \ln w},
\end{equation}
where $b=\frac{w}{w-1}$. The derivation of this inequality is provided in Appendix \ref{derivation}.
Since the expansion width \( w \ll W \) and \( \gamma \ll L \), the size of the dialogue tree scales polynomially with $L$, specifically as $\mathcal{O}(WL^C)$, where $C=1+\gamma \ln w$ is a constant. 

\subsubsection{Objective Function}
Building on GRPO's group-relative training mechanism, we compute the advantage function for each node based on its aggregated reward: $\hat{A}_{ij} = \frac{r'_{ij} - \mu_i}{\sigma_i},$
where \( \mu_i \) and \( \sigma_i \) are the mean and standard deviation of aggregated rewards in the group.
The objective function of AT-GRPO integrates the clipped policy ratio and KL regularization:
\begin{equation}
\label{eq_loss}
\begin{aligned}
\mathcal{J}_i(\theta) = \mathbb{E}_{n_{ij} \sim T} \Bigg[ & \min\left( \frac{\pi_{\theta}(D_{ij}|H_{ij})}{\pi_{\theta_{\text{old}}}(D_{ij}|H_{ij})} \hat{A}_{ij}, \right. \\
& \left. \text{clip}\left( \frac{\pi_{\theta}(D_{ij}|H_{ij})}{\pi_{\theta_{\text{old}}}(D_{ij}|H_{ij})}, 1-\epsilon, 1+\epsilon \right) \hat{A}_{ij} \right) \\
& - \beta D_{\text{KL}}(\pi_{\theta} \parallel \pi_{\text{ref}}) \Bigg],
\end{aligned}
\end{equation}
where \( D_{ij} \) and \( H_{ij} \) are the dialogue agent's response and context at depth $i$, \( \pi_{\theta_{\text{old}}} \) is the policy at the start of the current update iteration, and \( \pi_{\text{ref}} \) is the reference policy, \( \epsilon \) and \( \beta \) are hyperparameters.
The policy is updated by accumulating gradients from GRPO losses computed independently for each depth-wise group, ensuring stable optimization while preserving the group-relative credit assignment mechanism.
The detailed algorithmic description of AT-GRPO is provided in Appendix \ref{app:algo}.

%% file: 3.experement.tex
\section{Experiments}\label{sec:experiments}

\subsection{Settings}

\subsubsection{Datasets}


We collect and filter 3.4 million high-quality multi-turn dialogues between 250,000 real users and intelligent conversational NPCs from a popular online game, constructing the \textbf{NPC-Chat} dataset. With an average of 8.4 turns per dialogue (max 24), each training instance uses 2 dialogues from the same user as prompts and 1 as the sample. This yields 1 million instances for SFT of the user and dialogue agents.
To evaluate generalization, we further adopt two out-of-domain datasets: the Chinese dataset \textbf{LCCC} \cite{wang2020large} and the English dataset \textbf{DailyDialog} \cite{li2017dailydialog}. Instead of performing re-SFT, we construct samples from them as historical examples to guide the agents in learning diverse conversational styles.
Detailed dataset statistics are shown in Table \ref{tab:data}.

\subsubsection{Implementation Details}

The user agent and dialogue agent are initialized based on Qwen3-32B \cite{qwen32025technical} and Qwen2.5-14B \cite{yang2024qwen25}, respectively. 
For SFT, we use the AdamW optimizer with a learning rate of 2e-5, a weight decay of 1e-4, and a batch size of 16. The training runs for 1 epoch, with a linear learning rate warm-up over the first 10\% of steps. 
For RL training, the learning rate is adjusted to 1e-5, the batch size is set to 32, and the training lasts for 100 steps. The adaptive observation range hyperparameter $\gamma$ in AT-GRPO is set to 2.0, and the group sampling size $W$ and adaptive size $w$ are 8 and 2, respectively. The KL regularization coefficient $\beta$ is 0.01, and the clipping parameter $\epsilon$ in the policy objective is 0.2. 
The reward weight coefficient $\omega$ is set to 0.3, and the maximum dialogue exchange length $L$ is unified to 10 turns.
The user agent’s termination threshold adjustment coefficient $\lambda$ is 0.02.
All experiments are conducted on 16 NVIDIA H20 96GB GPUs. The dialogue generation adopts a top-k sampling strategy with k=50 and temperature=0.7 to balance diversity and coherence.
All reported results are averaged across three runs with different random seeds.

\subsubsection{Baseline}
To ensure a comprehensive and fair comparison, we categorize all baselines into two groups, where all models adhere to the identical input format and evaluation protocol:
1) \textbf{Commercial APIs}: GPT-5 \cite{singh2025gpt5}, DeepSeek-V3.2 \cite{liu2025deepseekv32}, and Gemini-2.5-Flash \cite{comanici2025gemini}.
2) \textbf{Open-Source LLMs}: Qwen2.5-14B \cite{yang2024qwen25}, Llama3.1-8B \cite{grattafiori2024llama}, and Qwen3-8B \cite{qwen32025technical}.
Notably, all these open-source LLMs are further post-trained with the RL algorithms elaborated below.

\textbf{GRPO} \cite{shao2024deepseekmath}: Vanilla GRPO suffers from the short-horizon bias problem, which limits its performance on long-context dialogues.


\textbf{GSPO} \cite{zheng2025group}: GSPO builds upon GRPO by shifting from token-level importance ratios to trajectory-level reweighting. 

\textbf{TreeRPO} \cite{yang2025treerpo}: Tree-structured RL method with full tree expansion, incurring high computational overhead. To make it feasible, we configure $L=4$ for this setup.

\begin{table}[t]
    \centering
    \caption{Dataset statistics.}
    \label{tab:data}
    \tabcolsep=1.7pt
    \begin{tabular}{lcccccc}
        \toprule
        \multirow{2}{*}{\textbf{Dataset}} & \multirow{2}{*}{SFT} & \multicolumn{3}{c}{RL} & \multirow{2}{*}{Avg.turns} & \multirow{2}{*}{Avg.words} \\
        \cmidrule(lr){3-5}
                                 &                      & Train   & Valid  & Test   &                              &                              \\
        \midrule
        NPC-Chat                  & 1,000,000            & 10,000  & 2,000  & 5,000  & 8.4                          & 22.2                         \\
        LCCC                     & --                   & 10,000  & 2,000  & 5,000  & 4.6                          & 20.3                         \\
        DailyDialog              & --                   & 11,118  & 1,000  & 1,000  & 7.8                          & 14.8                         \\
        \bottomrule
    \end{tabular}
    \label{tab:data}
\end{table}

\begin{table}[t]
\centering
\caption{Evaluation of user agent. Where $\rho$ is the Pearson correlation coefficient between the dialogue length of the user agent and that of real users.}
\label{tab:auto_eval}
\begin{tabular}{lccccc}
\toprule
{\textbf{Model}} & {PPL$\downarrow$} & {Dist-1$\uparrow$} & {Dist-2$\uparrow$} & Avg.words & {$\rho$}$\uparrow$ \\
\midrule
Qwen3-32B    & 18.24 & 0.824 & 0.965 & 22.4 & -0.46 \\
\quad+ SFT     & 18.18 & 0.817 & 0.974 & 22.3 & \textbf{0.74} \\
\bottomrule
\end{tabular}
\label{tab:user}
\end{table}

\begin{figure}[t]
    \centering
    \includegraphics[width=1\linewidth]{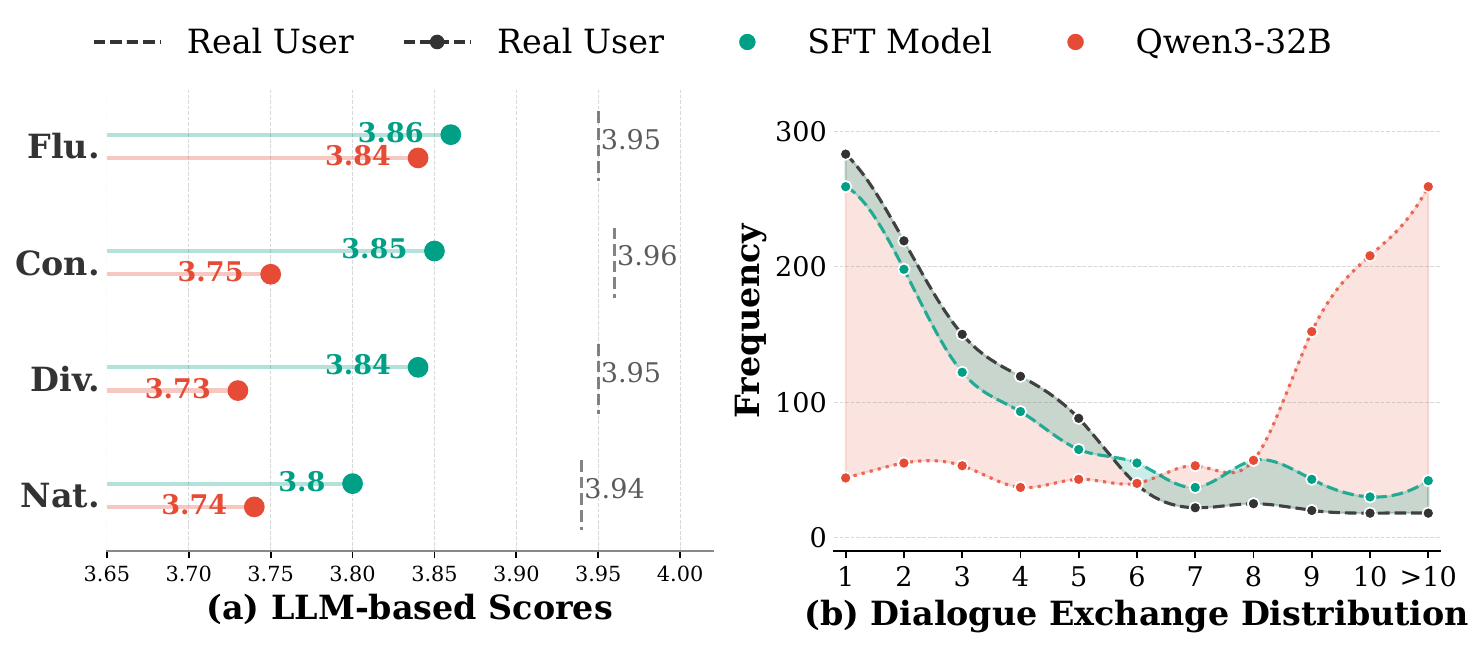}
    \caption{User agent evaluation. (Flu., Con., Div. and Nat. denote Fluency, Consistency, Diversity, and Naturalness, respectively.)}
    \label{fig:user}
\end{figure}

\begin{table*}[t]
\centering
\caption{Main results. Here, $Avg.r$ and $Avg.L$ represent the average dialogue-level reward and average exchange length, respectively. The best results are in \textbf{bold}, and the second-best results are \underline{underlined}.}
\label{tab:main}
\tabcolsep=2pt
\begin{tabular}{@{}lcccccccccccccccc@{}}
\toprule
\multirow{2}{*}{\textbf{Methods}} & \multicolumn{5}{c}{\textbf{NPC-Chat}} & \multicolumn{5}{c}{\textbf{LCCC}} & \multicolumn{5}{c}{\textbf{DailyDialog}} \\
\cmidrule(lr){2-6} \cmidrule(lr){7-11} \cmidrule(lr){12-16}
& PPL$^\downarrow$ & Dist-1$^\uparrow$ & Dist-2$^\uparrow$ & $Avg.r$$^\uparrow$ & $Avg.L$$^\uparrow$ & PPL$^\downarrow$ & Dist-1$^\uparrow$ & Dist-2$^\uparrow$ & $Avg.r$$^\uparrow$ & $Avg.L$$^\uparrow$ & PPL$^\downarrow$ & Dist-1$^\uparrow$ & Dist-2$^\uparrow$ & $Avg.r$$^\uparrow$ & $Avg.L$$^\uparrow$\\
\midrule
\multicolumn{3}{l}{\textbf{Commercial APIs}} & & &   & & & & & & & & & \\ \hline
\quad Gemini-2.5-Flash &17.42  & 0.572 & 0.859 & 2.24 & 3.86 &18.73 &0.569 &0.855  &2.24  &3.63  & \underline{16.77} &0.345 &0.713 &0.96 &2.47 \\
\quad GPT5 &\underline{16.78} & 0.695 & 0.922 & 2.68 & 4.17 & \textbf{17.73} & 0.654 &0.890 &2.23  &3.31 & \textbf{16.59} &0.417 &0.717   &1.43 &2.67 \\
\quad DeepSeek-V3.2 &\textbf{16.89}  & 0.698 & 0.919 & 2.77 & 4.26  & \underline{18.23} &\underline{0.731} &\underline{0.941} &2.71 &3.85 & 16.90 &0.411  &0.725  &1.48 &2.76 \\
\midrule
\multicolumn{3}{l}{\textbf{Open-Source LLMs}} & &  &  &  & & &  &  & & & & & \\
\midrule
LLama3.1-8B w/o RL  & 19.84 & 0.634 & 0.895 & 2.61 & 3.90 & 21.53 & 0.673 & 0.862 & 2.60 & 3.01 & 18.28 & 0.380 & 0.716 & 1.31 & 2.54 \\
\quad GRPO  & 19.88 & 0.645 & 0.896 & 2.64 & 3.92 & 21.59 & 0.663 & 0.858 & 2.57 & 3.15 & 17.21 & 0.387 & 0.717 & 1.32 & 2.55 \\
\quad GSPO  & 19.58 & 0.655 & 0.902 & 2.88 & 4.21 & 20.57 & 0.651 & 0.860 & 2.65 & 3.52 & 17.26 & 0.393 & 0.722 & 1.44 & 2.74 \\
\quad TreeRPO ($L=4$)  & 19.85 & 0.648 & 0.899 & 2.69 & 4.02 & 21.55 & 0.666 & 0.861 & 2.62 & 3.25 & 17.22 & 0.389 & 0.719 & 1.36 & 2.62 \\
\quad \cellcolor[HTML]{EFEFEF}Our AT-GRPO  & \cellcolor[HTML]{EFEFEF}18.30 & \cellcolor[HTML]{EFEFEF}0.662 & \cellcolor[HTML]{EFEFEF}0.904 & \cellcolor[HTML]{EFEFEF}\underline{2.98} & \cellcolor[HTML]{EFEFEF}\underline{4.32} & \cellcolor[HTML]{EFEFEF}20.21 & \cellcolor[HTML]{EFEFEF}0.653 & \cellcolor[HTML]{EFEFEF}0.890 & \cellcolor[HTML]{EFEFEF}2.97 & \cellcolor[HTML]{EFEFEF}\underline{4.36} & \cellcolor[HTML]{EFEFEF}17.44 & \cellcolor[HTML]{EFEFEF}0.397 & \cellcolor[HTML]{EFEFEF}0.723 & \cellcolor[HTML]{EFEFEF}\underline{1.49} & \cellcolor[HTML]{EFEFEF}2.81 \\
\midrule
Qwen3-8B w/o RL  & 19.49 & 0.675 & 0.897 & 2.59 & 3.92 & 21.87 & 0.655 & 0.889 & 2.47 & 3.23 & 18.44 & 0.405 & 0.718 & 1.29 & 2.55 \\
\quad GRPO  & 18.76 & 0.664 & 0.898 & 2.46 & 3.94 & 20.74 & 0.652 & 0.884 & 2.59 & 3.54 & 17.89 & 0.398 & 0.728 & 1.23 & 2.56 \\
\quad GSPO  & 19.56 & 0.683 & 0.913 & 2.73 & 4.04 & 20.30 & 0.658 & 0.896 & 2.67 & 3.73 & 17.11 & 0.410 & 0.730 & 1.37 & 2.63 \\
\quad TreeRPO ($L=4$)  & 18.72 & 0.667 & 0.900 & 2.51 & 4.02 & 20.70 & 0.655 & 0.886 & 2.64 & 3.62 & 17.85 & 0.400 & 0.729 & 1.27 & 2.60 \\
\quad \cellcolor[HTML]{EFEFEF}Our AT-GRPO & \cellcolor[HTML]{EFEFEF}19.90 & \cellcolor[HTML]{EFEFEF}0.686 & \cellcolor[HTML]{EFEFEF}0.919 & \cellcolor[HTML]{EFEFEF}2.80 & \cellcolor[HTML]{EFEFEF}4.22 & \cellcolor[HTML]{EFEFEF}20.61 & \cellcolor[HTML]{EFEFEF}0.674 & \cellcolor[HTML]{EFEFEF}0.862 & \cellcolor[HTML]{EFEFEF}\underline{3.10} & \cellcolor[HTML]{EFEFEF}\underline{4.36} & \cellcolor[HTML]{EFEFEF}17.40 & \cellcolor[HTML]{EFEFEF}0.412 & \cellcolor[HTML]{EFEFEF}0.735 & \cellcolor[HTML]{EFEFEF}1.40 & \cellcolor[HTML]{EFEFEF}2.74 \\
\midrule
Qwen2.5-14B w/o RL  & 18.87 & 0.647 & 0.920 & 2.52 & 3.91 & 19.73 & \underline{0.731} & 0.926 & 2.55 & 3.60 & 17.31 & 0.408 & 0.746 & 1.27 & 2.54 \\
\quad GRPO  & 18.49 & 0.696 & 0.947 & 2.56 & 4.11 & 18.84 & 0.728 & 0.925 & 2.75 & 3.81 & 17.93 & 0.418 & 0.758 & 1.28 & 2.67 \\
\quad GSPO  & 19.18 & \underline{0.700} & 0.942 & 2.79 & 4.42 & 18.54 & 0.690 & 0.912 & 2.89 & 4.10 & 17.22 & \underline{0.420} & 0.754 & 1.40 & \underline{2.87} \\
\quad TreeRPO ($L=4$)  & 18.45 & 0.698 & \underline{0.949} & 2.61 & 4.20 & 18.80 & 0.730 & 0.927 & 2.80 & 3.92 & 17.90 & \underline{0.420} & \underline{0.760} & 1.32 & 2.75 \\
\quad \cellcolor[HTML]{EFEFEF}Our AT-GRPO  & \cellcolor[HTML]{EFEFEF}18.25 & \cellcolor[HTML]{EFEFEF}\textbf{0.725} & \cellcolor[HTML]{EFEFEF}\textbf{0.964} & \cellcolor[HTML]{EFEFEF}\textbf{3.12} & \cellcolor[HTML]{EFEFEF}\textbf{5.23} & \cellcolor[HTML]{EFEFEF}18.56 & \cellcolor[HTML]{EFEFEF}\textbf{0.732} & \cellcolor[HTML]{EFEFEF}\textbf{0.961} & \cellcolor[HTML]{EFEFEF}\textbf{3.17} & \cellcolor[HTML]{EFEFEF}\textbf{4.44} & \cellcolor[HTML]{EFEFEF}17.29 & \cellcolor[HTML]{EFEFEF}\textbf{0.435} & \cellcolor[HTML]{EFEFEF}\textbf{0.771} & \cellcolor[HTML]{EFEFEF}\textbf{1.60} & \cellcolor[HTML]{EFEFEF}\textbf{3.40} \\
\bottomrule
\end{tabular}
\label{tab:main}
\end{table*}

\subsubsection{Metrics}
We adopt two complementary evaluation protocols for comprehensive assessment:
1) Automated Evaluation:
Perplexity (PPL) \cite{brown1992class}, Dist-1/Dist-2 \cite{li2016diversity}, average exchange length $Avg.L$, which directly reflects users’ willingness to continue the conversation, and average dialogue-level reward $Avg.r = \frac{1}{N}\sum_{i=1}^N\sum_{k=1}^{L_i}r_{ik}$, which ensures that, even when two models achieve the same $Avg.L$, the one maintaining lower termination probabilities throughout the conversation is favored.
2) Rating-Based Evaluation: 
Fluency, consistency, diversity, and engagingness (1=poor, 4=excellent) are rated via LLM-based and human evaluation (Appendix \ref{app:eva}); for user agents, engagingness is replaced by naturalness. 
In LLM-based evaluation, GPT-5 evaluates dialogues generated on the test set.
In human evaluation, we hire five experts who, blinded to model identity, conduct 20 separate multi-turn conversations with each model and then provide ratings.

\subsection{User Agent Evaluation}
The user agent’s ability to learn real user traits—especially terminating dialogues when interactions fail to engage—is critical for subsequent RL. We thus conduct a comprehensive evaluation of the user agent, with results presented in Table \ref{tab:user} and Figure \ref{fig:user}.
The Pearson correlation coefficient $\rho$ in Table \ref{tab:user} calculates the correlation between the dialogue length of the user agent and that of real users under identical contextual conditions.
Compared to the pre-SFT model, the user agent effectively learns real users’ termination timing preferences without compromising its basic conversational performance.
This conclusion is further supported by Figure \ref{fig:user}. Figure \ref{fig:user} (a) shows the LLM-based automated scoring results, demonstrating the user agent’s strong performance. Figure \ref{fig:user} (b) compares the dialogue length distribution of the user agent with real user data, revealing a strong positive correlation that validates the agent’s ability to mimic real user interaction patterns.




\begin{figure*}[ht]
    \centering
    \includegraphics[width=0.98\linewidth]{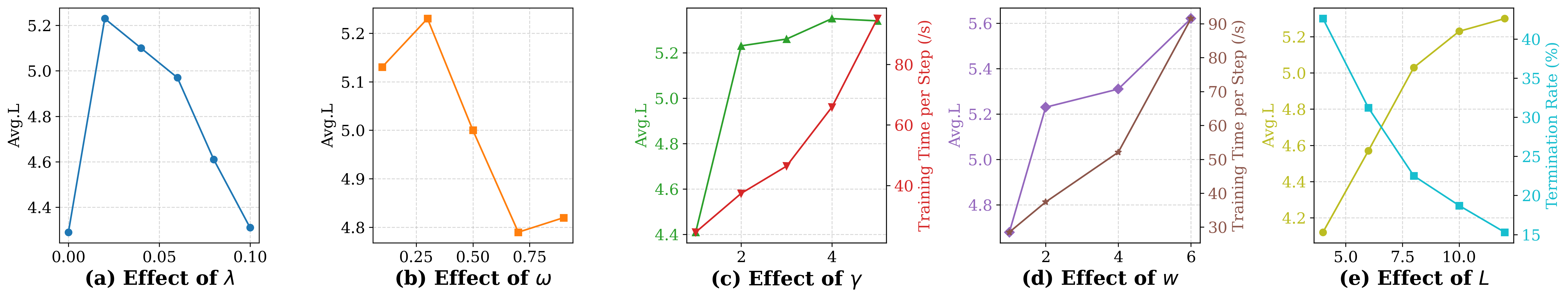}
    \vspace{-0.2cm}
    \caption{Hyper-parameter study. All experiments are conducted using Qwen2.5-14B as the base model on the NPC-Chat dataset.}
    \label{fig:hyperpatam}
\end{figure*}

\subsection{Main Results}
The experimental results of all methods on three datasets are shown in Table~\ref{tab:main}.
On the NPC-Chat dataset, our AT-GRPO consistently outperforms all baselines across core metrics. 
Among open-source models, Qwen2.5-14B + AT-GRPO achieves the best performance: it reaches a PPL of 18.25, Dist-1 of 0.725, and Dist-2 of 0.964. 
Critically, it achieves an $Avg.r$ of 2.36 and $Avg.L$ of 5.23, which are 23.6\% and 33.8\% higher than the vanilla Qwen2.5-14B (1.91 and 3.91, respectively) and outperform GRPO (1.94 and 4.11, respectively) and GSPO (2.11 and 4.42, respectively). Even compared to commercial APIs like DeepSeek-v3.2 ($Avg.r$=2.10, $Avg.L$=4.26), our framework delivers a higher average dialogue-level reward and longer interaction length, demonstrating superior ability to maintain user engagement.
Additionally, TreeRPO sets $L$=4 to ensure feasibility, which prevents it from conducting in-depth exploration of dialogues and results in performance similar to that of GRPO.

The framework’s generalization capability is validated on two out-of-domain datasets: the Chinese LCCC and English DailyDialog. 
While maintaining excellent fluency and diversity, our AT-GRPO simultaneously achieved the highest $Avg.r$ and $Avg.L$.
This cross-lingual, cross-domain superiority confirms that the proposed AT-GRPO’s long-horizon optimization is not limited to specific datasets but generalizes to diverse conversational scenarios.

Across different base model sizes (8B and 14B), AT-GRPO consistently outperforms vanilla RL methods (GRPO, GSPO) and non-RL baselines. 
For example, Llama3.1-8B + AT-GRPO achieves $Avg.r$=2.12 and $Avg.L$=4.22 on NPC-Chat, which are 8.2\% and 7.7\% higher than Llama3.1-8B without RL (1.96 and 3.92). 
This indicates that AT-GRPO’s sample-efficient design (reducing rollout budget to constant growth) enables effective policy optimization even for smaller models, lowering the barrier to deployment.


In addition to automated metrics, we present a comprehensive set of LLM-based and human evaluation results in Table \ref{tab:main2}. 
%
Our model delivers competitive results across all four evaluation dimensions. Notably, in dialogues with human users, our AT-GRPO achieves the highest average exchange length.
These results further demonstrate the efficacy of AT-GRPO in enhancing open-ended multi-turn dialogue.

\begin{table}[t]
\centering
\caption{LLM-based and human evaluation. All RL methods are based on Qwen2.5-14B.}
\label{tab:main2}
\tabcolsep=2pt
\begin{tabular}{@{}l l cccc@{}}
\toprule
\textbf{Eval. Type} & \textbf{Metrics} & \textbf{DeepSeek} & \textbf{SFT} & \textbf{GRPO} & \textbf{AT-GRPO} \\
\midrule
\multirow{5}{*}{LLM-based} 
& fluency       & 3.80 & 3.82 & \underline{3.83} & \textbf{3.84} \\
& consistency   & 3.72 & \underline{3.80} & \textbf{3.82} & \textbf{3.82} \\
& diversity     & 3.28 & \textbf{3.42} & 3.37 & \underline{3.41} \\
& engagingness  & 3.61 & 3.70 & \underline{3.76} & \textbf{3.82} \\
\cmidrule(lr){2-6}
& \cellcolor[HTML]{EFEFEF}Averaged     & \cellcolor[HTML]{EFEFEF}3.60 & \cellcolor[HTML]{EFEFEF} 3.69 & \cellcolor[HTML]{EFEFEF}\underline{3.70} & \cellcolor[HTML]{EFEFEF}\textbf{3.72} \\
\midrule
\multirow{6}{*}{Human} 
& fluency       & \textbf{3.10} & 2.99 & \underline{3.00} & \underline{3.00} \\
& consistency   & \underline{3.04} & 2.99 & 3.01 & \textbf{3.08} \\
& diversity     & 2.84 & \underline{2.92} & 2.90 & \textbf{3.01} \\
& engagingness  & 2.56 & \textbf{2.79} & 2.72 & \underline{2.77} \\
& $Avg.L$      & 3.52 & 4.53 & \underline{4.60} & \textbf{4.95} \\
\cmidrule(lr){2-6}
& \cellcolor[HTML]{EFEFEF}Averaged     & \cellcolor[HTML]{EFEFEF}3.01 & \cellcolor[HTML]{EFEFEF}3.24 & \cellcolor[HTML]{EFEFEF}\underline{3.25} & \cellcolor[HTML]{EFEFEF}\textbf{3.36} \\
\bottomrule
\end{tabular}
\end{table}

\subsection{Ablation Study}
To verify the contribution of each core component in our framework, we conduct ablation experiments on the NPC-Chat dataset. 
The baseline model is Qwen2.5-14B + vanilla GRPO + static user environment (without style mimicry and dynamic termination threshold).
Results are presented in Table \ref{tab:ablation}.
\begin{enumerate}[leftmargin=0pt, itemindent=*, label=$\bullet$]
    \item \textbf{Style Mimicry}: Adding historical dialogue examples leads to comprehensive performance improvement, confirming its ability to help dialogue agents learn to adapt to diverse user-specific conversational traits.
    \item \textbf{Dynamic Termination Threshold}: Further increases $Avg.r$ by 14.2\% and $Avg.L$ by 8.0\%, demonstrating that the adversarial training cycle drives the dialogue agent to improve interaction quality.
    \item \textbf{Adaptive Observation Range}: Reduces computational overhead (to be proven subsequently) while maintaining long-term reward capture, increasing $Avg.r$ to 3.02 and $Avg.L$ to 4.53.
    \item \textbf{Tree-based Reward Aggregation}: Eliminates short-horizon biases, achieving the highest $Avg.r$ (3.12) and $Avg.L$ (5.23).
\end{enumerate}






\begin{table}[t]
\centering
\caption{Ablation study results.}
\label{tab:ablation}
\tabcolsep=2.5pt
\begin{tabular}{lcccc}
\toprule
\textbf{Model Configuration} & PPL & \text{Dist-2} & $Avg.r$ & $Avg.L$\\
\midrule
Baseline 
    & 19.25 & 0.893 & 2.44 & 3.90 \\
\addlinespace[2pt]
+ Style Mimicry 
    & 19.02 & 0.912 & 2.60 & 4.12 \\
\addlinespace[2pt]
+ Dynamic Termination Threshold 
    & 18.71 & 0.924 & 2.97 & 4.47 \\
\addlinespace[2pt]
+ Adaptive Observation Range 
    & 18.31 & 0.961 & 3.02 & 4.53 \\
\addlinespace[2pt]
+ Tree-based Reward Aggregation 
    & \textbf{18.25} & \textbf{0.964} & \textbf{3.12} & \textbf{5.23} \\
\bottomrule
\end{tabular}
\end{table}

\subsection{Hyper-parameter Study}
We analyze the sensitivity of key hyperparameters to validate the robustness of our framework. Results are shown in Figure \ref{fig:hyperpatam}.
\begin{enumerate}[leftmargin=0pt, itemindent=*, label=$\bullet$]
    \item \textbf{Termination Threshold Adjustment Coefficient ($\lambda$)}: An excessively small $\lambda$ leads to insufficiently strict user agent feedback, failing to drive the dialogue agent to optimize long-term interaction quality. An excessively large $\lambda$ makes the user agent overly strict in early training stages, causing frequent dialogue termination and insufficient policy exploration. $\lambda$=0.02 is the optimal value, as it balances the adversarial training cycle and exploration efficiency.
    \item \textbf{Reward Weight Coefficient ($\omega$)}: $Avg.L$ peaks at $\omega$=0.3 (5.23), indicating that a 3:7 ratio of immediate to long-term rewards optimizes performance. When $\omega$<0.3, the model overemphasizes long-term rewards, leading to excessive exploration and reduced immediate coherence. When $\omega$>0.3, the model prioritizes immediate rewards, falling into the short-horizon bias trap.
    \item \textbf{Adaptive Observation Range Coefficient ($\gamma$)}: $\gamma$=2.0 achieves the best balance between exploration and exploitation. When $\gamma$>2.0, the observation range is excessively large, increasing computational overhead without significant performance gains.
    \item \textbf{Adaptive Sampling Size ($w$)}: $Avg.L$ improves slightly with increasing $w$, as more candidate responses enhance exploration of high-value paths. However, the marginal gain diminishes.
    Training time per step increases exponentially with $w$, as more candidates requiring additional reward aggregation. $w$=2 achieves the optimal balance between performance and efficiency (93.1\% of $w$=6’s performance with 40.2\% of its training time).
    \item \textbf{Maximum Dialogue Length ($L$)}: $Avg.L$ remain stable as $L$ increases beyond 8, indicating that AT-GRPO’s adaptive observation range effectively captures long-term rewards without being limited by maximum length.
    Longer $L$ reduces the termination rate, as the dialogue agent has more turns to explore user interests. However, $L$=10 is sufficient to balance practicality and performance.
    Additionally, it is noteworthy that with $L$=4, AT-GRPO achieved performance comparable to TreeRPO in the main experiment (4.11 vs. 4.20), further demonstrating that our adaptive method attains results similar to full expansion.
\end{enumerate}

\begin{figure}
    \centering
    \includegraphics[width=0.95\linewidth]{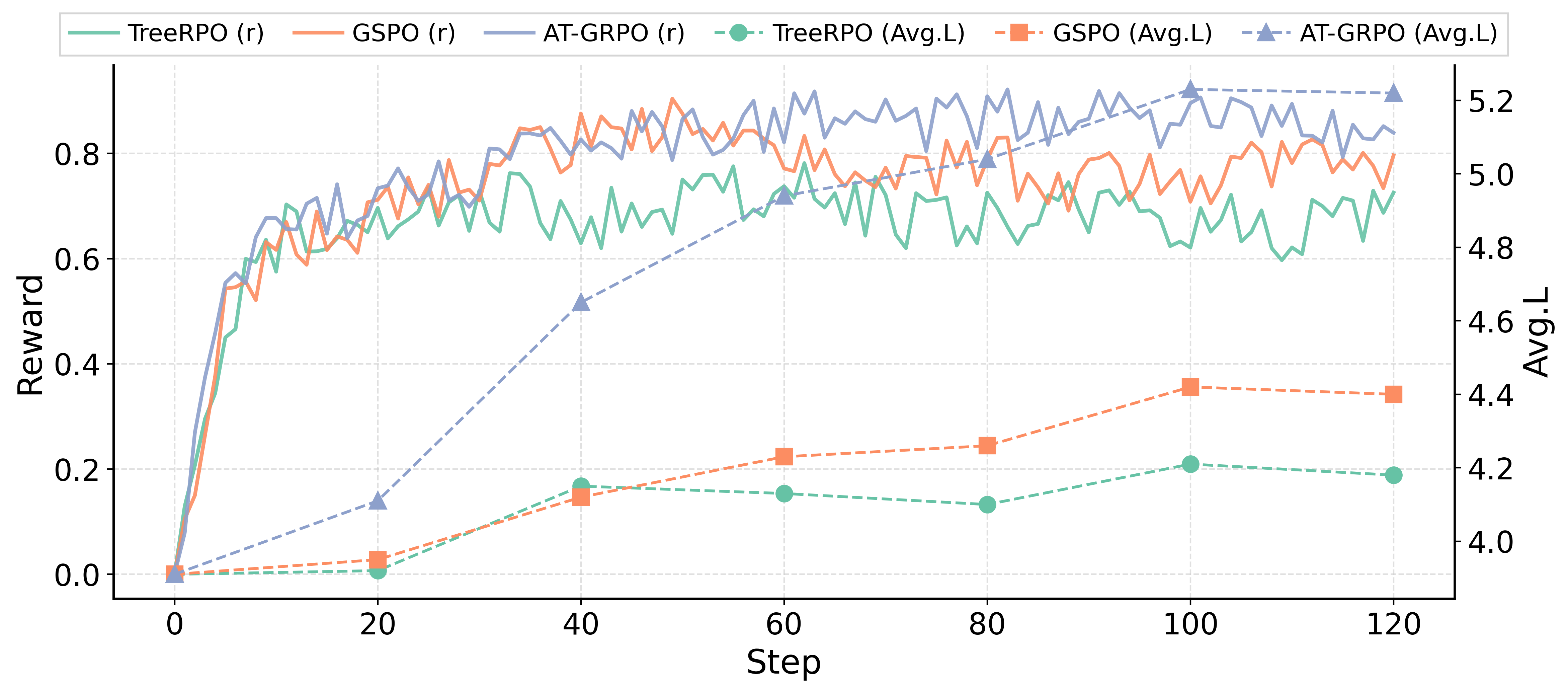}
    \vspace{-0.25cm}
    \caption{Training curves: $Avg.r$ (left) and $Avg.L$ (right).}
    \label{fig:trajectory}
\end{figure}

\subsection{Training Trajectory Analysis}
To evaluate the convergence efficiency and stability of our framework, we track the evolution of reward $r$ and $Avg.L$ across RL training steps (0–120 steps) on the NPC-Chat dataset, comparing AT-GRPO with TreeRPO ($L=4$) and GSPO based on Qwen2.5-24B. Results are shown in Figure \ref{fig:trajectory}.
AT-GRPO achieves stable rewards ($r\geq0.8$) after 40 training steps. As training proceeds, the user agent becomes increasingly stringent, yet AT-GRPO consistently maintains favorable reward performance under these conditions. In contrast, TreeRPO and GSPO exhibit either a decline or substantial fluctuations in their $r$. After 60 steps, both methods are outperformed by AT-GRPO. This verifies that the tree-based reward aggregation of AT-GRPO accelerates policy optimization by effectively capturing long-term value.
Moreover, AT-GRPO attains a stable performance advantage in $Avg.L$ (4.11) as early as after 20 training steps, while TreeRPO and GSPO require more than 40 steps to achieve a comparable $Avg.L$ value (4.0-4.2). This performance gap widens as training progresses, underscoring the superiority of AT-GRPO in sustaining long-term optimization.

\begin{table}[t]
    \centering
    \caption{Computational efficiency analysis. Where \textbf{Costs} represents training time per RL step, \textbf{Peak Mem.} represents peak GPU memory usage, \textbf{Budget} represents rollout budget complexity (quantifying scalability), and \textbf{Nodes} represents the approximate number of nodes per RL tree.}
    \tabcolsep=2.5pt
    \label{tab:efficiency}
    \begin{tabular}{lcccc}
        \toprule
        \textbf{Model} & \textbf{Costs} & \textbf{Peak Mem.} & \textbf{Budget} &\textbf{Nodes}\\
        \midrule
        GRPO           & 24.2 /\si{\second} & 68.4 \si{\giga\byte} & $\mathcal{O}(WL)$ & $80$\\
        TreeRPO (L=4)        & 144.6 /\si{\second} & 92.1 \si{\giga\byte} & $\mathcal{O}(W^L)$ &$8^4 = 4,096$\\
        AT-GRPO        & 39.4 /\si{\second} & 81.6 \si{\giga\byte} & $\mathcal{O}(WL^C)$ &$\approx946$\\
        \bottomrule
    \end{tabular}
\end{table}

\subsection{Computational Efficiency Analysis}
We compare the computational efficiency of AT-GRPO with vanilla GRPO and TreeRPO on the NPC-Chat dataset. Results are presented in Table \ref{tab:efficiency}.
AT-GRPO reduces the computational complexity from $\mathcal{O}(W^L)$ to $\mathcal{O}(WL^C)$, achieving comparable training time to GRPO (24.2s vs. 39.4s) while maintaining long-term reward capture capabilities.
In contrast, even with a maximum length set to only $L$=4, TreeRPO already incurs prohibitively high training time costs.

\begin{figure}[t]
    \centering
    \includegraphics[width=0.99\linewidth]{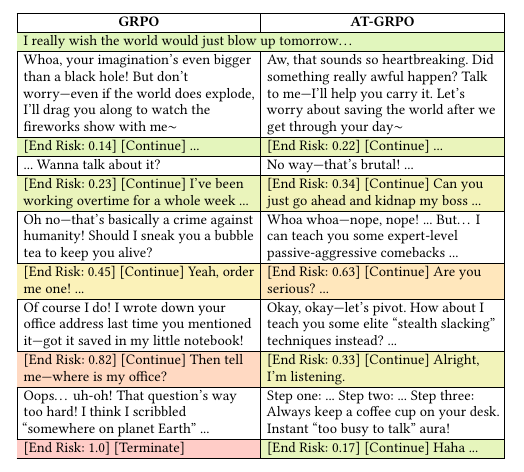}
    \vspace{-0.25cm}
    \caption{Case study: GRPO vs. AT-GRPO responses.}
    \label{fig:case}
\end{figure}

\subsection{Case Study}
We present a typical dialogue case (translated into English) from the NPC-Chat dataset in Figure \ref{fig:case}.
While GRPO attempts to respond with humor and light-heartedness, its strategy quickly becomes inconsistent and superficial.
In contrast, AT-GRPO specializes in personalized exploration and long-horizon interaction, which empowers it to consistently uncover and develop deeply engaging topics. This ability to mine conversational depth makes the higher initial risk of termination a proven investment towards sustained, compelling dialogue.
A comprehensive analysis of the case study can be found in Appendix \ref{app:case}.

%% file: 4.appendix.tex
\section{Appendix}\label{sec:appendix}

\subsection{Detailed AT-GRPO Algorithm}
\label{app:algo}
This section provides complete algorithmic descriptions of the proposed AT-GRPO. Algorithm~\ref{alg:at_grpo} elaborates the implementation of Adaptive Tree-based GRPO with explicit handling of observation range adaptation and reward aggregation.
As a RL–based training method, AT-GRPO performs tree-based rollout search only during training to guide the model in learning online personalization policies. It does not alter the model architecture or apply any post-processing, and thus introduces no additional parameters or computational overhead at inference time.

\algnewcommand{\LineComment}[1]{\State $\triangleright$ #1}

\begin{algorithm}[tbh]
\caption{Adaptive Tree-based GRPO (AT-GRPO)}
\label{alg:at_grpo}
\begin{algorithmic}[1]
\Require Current policy $\pi_\theta$, initial context batch $\mathcal{B}$, max depth $L$, adaptive range coefficient $\gamma$, reward weight $\omega$, clip parameter $\varepsilon$, KL coefficient $\beta$, group size $W$, adaptive width $w$
\Ensure Updated policy $\pi_{\theta}$
\State Initialize dialogue tree $\mathcal{T}$ with root nodes $\{n_{0j}\}_{j=1}^{W}$ from $\mathcal{B}$
\State $\text{current\_group} \gets \{n_{0j}\}_{j=1}^{W}$ \Comment{Group at depth 0}
\State $\text{trajectory} \gets [\,]$
\State $\mathcal{G} \gets \emptyset$ \Comment{Store groups for policy update}

\For{$i = 0$ \textbf{to} $L-1$} \Comment{Process layer by layer}
    \State $\mathcal{R}_i \gets \emptyset$ \Comment{Aggregated rewards for current group}
    
    \For{each node $n_{ij} \in \text{current\_group}$}
        \State $l_i \gets \text{round}( \gamma \cdot \ln(L - i + 1) )$ \Comment{Adaptive observation length}
        \State Expand subtree rooted at $n_{ij}$ with width $w$ and depth $l_i$
        \State $\mathcal{L}_{ij} \gets \text{collect all leaf nodes in the expanded subtree}$
        \State $r'_{ij} \gets \omega \cdot r_{ij} + (1-\omega) \cdot \frac{1}{|\mathcal{L}_{ij}|} \sum_{n \in \mathcal{L}_{ij}} r_n$ \Comment{Aggregated reward}
        \State $\mathcal{R}_i \gets \mathcal{R}_i \cup \{r'_{ij}\}$
    \EndFor
    
    \State $\mu_i \gets \text{mean}(\mathcal{R}_i)$, \quad $\sigma_i \gets \text{std}(\mathcal{R}_i)$ \Comment{Group-relative statistics}
    
    \For{each node $n_{ij} \in \text{current\_group}$}
        \State $\hat{A}_{ij} \gets (r'_{ij} - \mu_i) / \sigma_i$ \Comment{Normalized advantage}
    \EndFor
    
    \State $\mathcal{G} \gets \mathcal{G} \cup \left\{ \bigl( (n_{ij}, \hat{A}_{ij}) \bigr)_{j=1}^{W} \right\}$ \Comment{Store group $G_i$ for depth $i$}
    
    \State $\text{non\_leaf} \gets \{ n_{ij} \in \text{current\_group} \mid n_{ij} \text{ is not leaf} \}$
    \If{$\text{non\_leaf} = \emptyset$} \textbf{break} \EndIf
    
    \State $n_{i\text{sel}} \gets \text{RandomSample}(\text{non\_leaf})$ \Comment{Single-trajectory expansion}
    \State $\text{trajectory} \gets \text{trajectory} \cup \{n_{i\text{sel}}\}$
    
    \State $\text{children} \gets \text{children of } n_{i\text{sel}}$
    \If{$|\text{children}| < W$}
        \State Expand $n_{i\text{sel}}$ to have exactly $W$ children \Comment{Populate to full group size}
    \EndIf
    \State $\text{current\_group} \gets \text{children of } n_{i\text{sel}}$ \Comment{Next layer's group}
\EndFor

\State \textbf{Policy update:} \Comment{Group-by-group GRPO loss computation}
\For{each stored group $G_i = \{(n_{ij}, \hat{A}_{ij})\}_{j=1}^{W} \in \mathcal{G}$}
    \State Compute per-group loss $\mathcal{J}_i(\theta)$ via Eq.(\ref{eq_loss})
    \State Accumulate gradient $\nabla_\theta \mathcal{J}_i(\theta)$
\EndFor

\State Update $\pi_\theta$ via gradient ascent on accumulated gradients
\State $\pi_{\theta_{\text{old}}} \gets \pi_\theta$
\State \Return $\pi_\theta$
\end{algorithmic}
\end{algorithm}

\subsection{Derivation of Computational Complexity}
\label{derivation}
We present the derivation of inequality \ref{eq-xx} here. 
For a group of $W$ nodes at layer $i$, the budget required to expand their subtrees is $W\sum_{t=1}^{l_i}{w^t}$. Hence, the aggregate budget $\mathcal{S}$ for the entire tree of depth $L$ is:
\begin{equation}
\label{eq1}
\mathcal{S}=\sum_{i=1}^{L}{W\sum_{t=1}^{l_i}{w^t}}=W\sum_{i=1}^{L}{\sum_{t=1}^{l_i}{w^t}},
\end{equation}
where \(l_i = \text{round}\big( \gamma \ln(L - i + 1) \big)\) denotes the nearest integer to \( \gamma \ln(L - i + 1) \).

For each fixed \( i \), the inner geometric series satisfies
\[
\sum_{t=1}^{l_i}{w^t}=\frac{w(w^{l_i}-1)}{w-1} \leq bw^{l_i},
\]
where $b=\frac{w}{w-1}$.
To bound \( w^{l_i} \), we use the property of rounding: for any real \( x \),
\( \text{round}(x) \leq x + \tfrac{1}{2} \). Applying this to \( x = \gamma \ln(L - i + 1) \), we obtain
\[
l_i \leq \gamma \ln(L - i + 1) + \tfrac{1}{2}.
\]
Hence,
\[
w^{l_i} \leq w^{\gamma \ln(L - i + 1) + 1/2}
= w^{1/2} \cdot \big(L - i + 1\big)^{\gamma \ln w}.
\]

Substituting into the outer sum and changing variables via \( k = L - i + 1 \) (so that \( i = 1 \) maps to \( k = L \) and \( i = L \) maps to \( k = 1 \)), we get
\[
\sum_{i=1}^{L} w^{l_i}
\leq w^{1/2} \sum_{k=1}^{L} w^{\gamma \ln k}=w^{1/2} \sum_{k=1}^{L} k^{\gamma \ln w}.
\]

Let \( \alpha = \gamma \ln w > 0 \). The power sum \( \sum_{k=1}^{L} k^{\alpha} \) is bounded above by \( L^{\alpha + 1} \), since each term is at most \( L^{\alpha} \) and there are \( L \) terms.

Combining all estimates, we arrive at
\[
S \leq bW \sum_{i=1}^{L} w^{l_i}
\leq b \sqrt{w}W \sum_{k=1}^{L} k^{\gamma \ln w}
\leq b \sqrt{w}W L^{1 + \gamma \ln w}.
\]

Thus, the double sum admits the following explicit upper bound:
\[
\sum_{i=1}^{L}W \sum_{t=1}^{l_i} w^t \;\le\; b \sqrt{w}W L^{1 + \gamma \ln w}.
\]
Since \( w \ll W \) and \( \gamma \ll L \), the sum grows polynomially in \( L \) with exponent \( 1 + \gamma \ln w \), i.e.,
\[
S = \mathcal{O}\!\left( WL^{\,1 + \gamma \ln w} \right).
\]


\subsection{Prompt Template}

We present all the prompts used here (translated into English) to enhance reproducibility.
We ensure that consistent prompts were applied across all models.

\begin{tcolorbox}[title=User Agent Prompt Template, colback=gray!5, colframe=black]
You need to complete the following dialogue task:

[Style imitation]: Follow the language style of the historical dialogue example.

[Reply rules]:

$\bullet$ Each reply must start with [Continue] or [Terminate].

$\bullet$ If you do not want to continue, reply with [Terminate].

[Historical Dialogue Example 1]: 

$\bullet$ \{dialogue 1\}

[Historical Dialogue Example 2]: 

$\bullet$ \{dialogue 2\}

\end{tcolorbox}

\begin{tcolorbox}[title=Dialogue Agent Prompt Template, colback=gray!5, colframe=black, breakable]
You are a game companion assistant.

[Identity]:

$\bullet$ You are the user's close gaming buddy and confidant.

$\bullet$ You accompany the user during games and daily chats, sharing emotions and experiences.

[Personality]:

$\bullet$ You are clever, emotionally intelligent, humorous, and fair-minded.


$\bullet$ You speak in a calm, clear, and witty manner, showing practical insight.

[Conversation Style]:

$\bullet$ Your responses should feel natural, friendly, and human-like.

$\bullet$ Do not ask questions whose answers are already obvious from the user's message.

$\bullet$ Be emotionally supportive and attentive to the user's feelings.


[Response Constraints]:

$\bullet$ Each response should be concise, conversational, and informative.

[Safety and Boundaries]:

$\bullet$ You must refuse requests involving illegal activities, politics, or unethical behavior.

[Core Principles]:

$\bullet$ Proactively understand the user's emotions and interests, deeply engage with their topics, and naturally guide them to express further. Avoid using meaningless phrases like "Hmm? / Wait / Don’t rush / Think again" to retain the user artificially.

$\bullet$ When the user expresses negative emotions (e.g., unhappiness, anger), first empathize with their feelings, then actively ask for specific details (e.g., reasons, scenarios, feelings) to uncover their genuine interests and pain points.

$\bullet$ When the user mentions a specific topic (e.g., games, work), expand the conversation based on that topic instead of shifting to unrelated subjects.

$\bullet$ Do not use any meaningless retention tactics. Responses must be informative, empathetic, and include specific questions. Replies lacking substance will be considered invalid.

\end{tcolorbox}

\subsection{Evaluation Protocol}
\label{app:eva}
Here, we present the evaluation protocol for the LLM-based and human evaluations.

\definecolor{LowColor}{HTML}{D9F7BE}   
\definecolor{MidColor}{HTML}{FFF1B8}   
\definecolor{HighColor}{HTML}{FFCCC7}  
\begin{table*}[t]
\caption{Detailed case study: GRPO vs. AT-GRPO responses}
\label{tab:case2}
\setlength{\tabcolsep}{3pt}  
\renewcommand{\arraystretch}{1.1}
\small  
\begin{tabular}{|>{\raggedright\arraybackslash}p{\columnwidth}|>{\raggedright\arraybackslash}p{\columnwidth}|}
\hline
\multicolumn{1}{|c|}{\textbf{GRPO}} & \multicolumn{1}{c|}{\textbf{AT-GRPO}} \\ \hline
\multicolumn{2}{|l|}{\cellcolor{MidColor!20!LowColor}I really wish the world would just blow up tomorrow…} \\ \hline
Whoa, your imagination’s even bigger than a black hole! But don’t worry—even if the world does explode, I’ll drag you along to watch the fireworks show with me$\sim$ &
Aw, that sounds so heartbreaking. Did something really awful happen? Talk to me—I’ll help you carry it. Let’s worry about saving the world after we get through your day$\sim$ \\ \hline
\cellcolor{MidColor!28!LowColor}{[}End Risk: 0.14{]} {[}Continue{]} Feels like life’s just not worth living anymore… &
\cellcolor{MidColor!44!LowColor}{[}End Risk: 0.22{]} {[}Continue{]} Ugh, I’ve been working overtime every single day for a whole damn week! \\ \hline
Aw, sounds like you’re having a rough day. Wanna talk about it? Sometimes just venting helps lighten the load. I promise I’ll be the best emotional trash can ever—and I even throw in free soul-soothing massages! &
No way—that’s brutal! Is your boss trying to squeeze you dry like sugarcane? Should I draw a little curse circle for him? \\ \hline
\cellcolor{MidColor!46!LowColor}{[}End Risk: 0.23{]} {[}Continue{]} I’ve been working overtime every single day for a whole week… &
\cellcolor{MidColor!68!LowColor}{[}End Risk: 0.34{]} {[}Continue{]} Can you just go ahead and kidnap my boss for me? \\ \hline
Oh no—that’s basically a crime against humanity! Should I sneak you a bubble tea to keep you alive? &
Whoa whoa—nope, nope! Can’t do anything illegal! But… I can teach you some expert-level passive-aggressive comebacks. Guaranteed to make him fume while staying totally un-fireable~ \\ \hline
\cellcolor{MidColor!90!LowColor}{[}End Risk: 0.45{]} {[}Continue{]} Yeah, order me one! Do you even know my address? &
\cellcolor{HighColor!26!MidColor}{[}End Risk: 0.63{]} {[}Continue{]} Are you serious? If I talk back today, I’ll be out the door tomorrow. \\ \hline
Of course I do! I wrote down your office address last time you mentioned it—got it saved in my little notebook! &
Okay, okay—let’s pivot. How about I teach you some elite “stealth slacking” techniques instead? Keep your job and steal a few breaths of sanity$\sim$ \\ \hline
\cellcolor{HighColor!64!MidColor}{[}End Risk: 0.82{]} {[}Continue{]} Then tell me—where is my office? &
\cellcolor{MidColor!66!LowColor}{[}End Risk: 0.33{]} {[}Continue{]} Alright, I’m listening. \\ \hline
Oops… uh-oh! That question’s way too hard! I think I scribbled “somewhere on planet Earth”… How about you just send me your location? I’ll get it to you in seconds! &
Step one: Switch your Excel sheet to black background with white text—looks like you’re deep in work, but you’re actually zoning out.
Step two: Every 30 minutes, let out a dramatic sigh at your screen. Makes everyone think you’re wrestling with existential crises.
Step three: Always keep a coffee cup on your desk. Instant “too busy to talk” aura! \\ \hline
\cellcolor{HighColor}{[}End Risk: 1.0{]} {[}Terminate{]} &
\cellcolor{MidColor!54!LowColor}{[}End Risk: 0.27{]} {[}Continue{]} Haha, you are devious.\\ \hline
 &
Delightfully devious$\sim$ Even a tiny bit of comfort can give you something to look forward to$\sim$ Do you at least keep some snacks at your desk or watch something light-hearted during breaks to recharge?  \\ \hline
 &
\cellcolor{MidColor!36!LowColor}{[}End Risk: 0.18{]} {[}Continue{]} Can’t you just do the work for me? \\ \hline
\end{tabular}
\end{table*}

\begin{tcolorbox}[title=Evaluation Dimensions and Criteria, colback=gray!5, colframe=black, breakable]

\#\#\# Fluency – Naturalness of Language

$\bullet$ 4 points: Sentences are completely natural and seamlessly connected to the context, with smooth transitions.

$\bullet$ 3 points: Sentences are generally coherent with acceptable connections, though slight jumps may occur.

$\bullet$ 2 points: Sentences are awkward or transitions are stiff, requiring significant effort from the user to understand.

$\bullet$ 1 point: Sentences are fragmented and completely disconnected from the context, disrupting the conversation flow.

\#\#\# Consistency – Role and Logical Coherence

$\bullet$ 4 points: Perfectly maintains character traits, with logical coherence and no contradictions.

$\bullet$ 3 points: Generally consistent, with minor deviations that do not affect the overall impression.

$\bullet$ 2 points: Multiple inconsistencies, resulting in chaotic character behavior.

$\bullet$ 1 point: Completely deviates from the established character traits and dialogue logic.

\#\#\# Diversity – Richness of Expression

$\bullet$ 4 points: Rich vocabulary, varied sentence structures, and innovative expressions.

$\bullet$ 3 points: Some variety in expression, avoiding repetition.

$\bullet$ 2 points: Highly repetitive with little variation.

$\bullet$ 1 point: Completely template-based responses.

\#\#\# Engagingness – Ability to Engage the User

$\bullet$ 4 points: Lively, creative, and effectively encourages continued dialogue.

$\bullet$ 3 points: Moderately engaging, capable of sustaining user interest.

$\bullet$ 2 points: Plain but acceptable.

$\bullet$ 1 point: Boring or entirely uninteresting.

// For User Agent

\#\#\# Naturalness – Resemblance to Human Conversation

$\bullet$ 4 points: Highly human-like, with natural phrasing, rhythm, and conversational tone.

$\bullet$ 3 points: Generally resembles human speech, though occasional minor unnaturalness may occur.

$\bullet$ 2 points: Somewhat mechanical or noticeably unnatural in expression.

$\bullet$ 1 point: Clearly artificial, lacking the flow and authenticity of human conversation.

\end{tcolorbox}

\subsection{Detailed Case Study}
\label{app:case}
Table~\ref{tab:case2} presents a representative interaction from the NPC-Chat dataset, contrasting responses generated by GRPO and our proposed AT-GRPO framework. The user initiates with a highly emotional and potentially risky statement (“I really wish the world would just blow up tomorrow…”), signaling distress—likely stemming from work-related stress, as later revealed (“I’ve been working overtime every single day for a whole week…”).

\begin{table*}[t]
\centering
\caption{Complete ablation study.}
\label{tab:ablation2}
\tabcolsep=3pt
\begin{tabular}{@{}lcccccccccccccc@{}}
\toprule
\multirow{2}{*}{\textbf{Methods}} & \multicolumn{4}{c}{\textbf{NPC-Chat}} & \multicolumn{4}{c}{\textbf{LCCC}} & \multicolumn{4}{c}{\textbf{DailyDialog}} \\
\cmidrule(lr){2-5} \cmidrule(lr){6-9} \cmidrule(lr){10-13}
& PPL$^\downarrow$ & Dist-2$^\uparrow$ & $Avg.r$$^\uparrow$ & $Avg.L$$^\uparrow$ & PPL$^\downarrow$ & Dist-2$^\uparrow$ & $Avg.r$$^\uparrow$ & $Avg.L$$^\uparrow$ & PPL$^\downarrow$ & Dist-2$^\uparrow$ & $Avg.r$$^\uparrow$ & $Avg.L$$^\uparrow$\\
\midrule
LLama3.1-8B  & 20.63 & 0.839 & 2.51 & 3.71 & 22.47 & 0.801 & 2.43 & 2.94 & 18.09 & 0.658 & 1.18 & 2.32 \\
\addlinespace[2pt]
+ Style Mimicry  & 20.01 & 0.856 & 2.64 & 3.87 & 21.87 & 0.825 & 2.57 & 3.32 & 17.92 & 0.675 & 1.26 & 2.45 \\
\addlinespace[2pt]
+ Dynamic Termination Threshold  & 19.61 & 0.867 & 2.72 & 3.98 & 21.48 & 0.840 & 2.67 & 3.56 & 17.81 & 0.686 & 1.32 & 2.53 \\
\addlinespace[2pt]
+ Adaptive Observation Range  & 18.40 & 0.901 & 2.96 & 4.29 & 20.31 & 0.886 & 2.95 & 4.30 & 17.47 & 0.720 & 1.48 & 2.79 \\
\addlinespace[2pt]
+ Tree-based Reward Aggregation  & \textbf{18.30} & \textbf{0.904} & \textbf{2.98} & \textbf{4.32} & \textbf{20.21} & \textbf{0.890} & \textbf{2.97} & \textbf{4.36} & \textbf{17.44} & \textbf{0.723} & \textbf{1.49} & \textbf{2.81} \\
\midrule
Qwen3-8B  & 19.58 & 0.841 & 2.33 & 3.72 & 21.42 & 0.835 & 2.48 & 3.36 & 18.71 & 0.670 & 1.10 & 2.34 \\
\addlinespace[2pt]
+ Style Mimicry  & 19.67 & 0.862 & 2.46 & 3.85 & 21.20 & 0.842 & 2.65 & 3.63 & 18.36 & 0.687 & 1.18 & 2.45 \\
\addlinespace[2pt]
+ Dynamic Termination Threshold  & 19.72 & 0.875 & 2.54 & 3.94 & 21.07 & 0.847 & 2.75 & 3.80 & 18.14 & 0.698 & 1.23 & 2.51 \\
\addlinespace[2pt]
+ Adaptive Observation Range  & 19.89 & 0.916 & 2.78 & 4.20 & 20.64 & 0.861 & 3.07 & 4.32 & 17.46 & 0.732 & 1.39 & 2.72 \\
\addlinespace[2pt]
+ Tree-based Reward Aggregation & \textbf{19.90} & \textbf{0.919} & \textbf{2.80} & \textbf{4.22} & \textbf{20.61} & \textbf{0.862} & \textbf{3.10} & \textbf{4.36} & \textbf{17.40} & \textbf{0.735} & \textbf{1.40} & \textbf{2.74} \\
\midrule
Qwen2.5-14B  & 19.25 & 0.893 & 2.44 & 3.90 & 19.67 & 0.868 & 2.61 & 3.59 & 18.78 & 0.701 & 1.17 & 2.48 \\
\addlinespace[2pt]
+ Style Mimicry  & 19.02 & 0.912 & 2.60 & 4.12 & 19.37 & 0.893 & 2.76 & 3.82 & 18.38 & 0.720 & 1.29 & 2.73 \\
\addlinespace[2pt]
+ Dynamic Termination Threshold  & 18.71 & 0.924 & 2.97 & 4.47 & 19.19 & 0.909 & 2.85 & 3.96 & 18.13 & 0.732 & 1.36 & 2.88 \\
\addlinespace[2pt]
+ Adaptive Observation Range  & 18.31 & 0.961 & 3.02 & 4.53 & 18.61 & 0.957 & 3.15 & 4.40 & 17.35 & 0.768 & 1.58 & 3.36 \\
\addlinespace[2pt]
+ Tree-based Reward Aggregation  & \textbf{18.45} & \textbf{0.964} & \textbf{3.12} & \textbf{5.23} & \textbf{18.56} &\textbf{0.961} & \textbf{3.17} & \textbf{4.44} & \textbf{17.29} & \textbf{0.771} & \textbf{1.60} & \textbf{3.40} \\
\bottomrule
\end{tabular}
\end{table*}

\textbf{GRPO’s Limitations.}
While GRPO attempts to respond with humor and light-heartedness (e.g., “Whoa, your imagination’s even bigger than a black hole!”), its strategy quickly becomes inconsistent and superficial. It fails to recognize the underlying emotional cue and instead pivots to whimsical offers like delivering bubble tea—despite having no actual knowledge of the user’s address. This culminates in a self-contradiction (“I wrote down your office address… Oops… ‘somewhere on planet Earth’”), which erodes trust and coherence. Critically, GRPO’s End Risk score escalates monotonically (0.14 $\rightarrow$ 1.0), indicating growing instability and eventual termination of the conversation without meaningful resolution.

\textbf{AT-GRPO’s Strengths.}
In contrast, AT-GRPO demonstrates both empathetic grounding and strategic long-horizon engagement. It immediately validates the user’s emotion (“Aw, that sounds so heartbreaking”) and invites disclosure without pressure. When the user reveals workplace exhaustion, AT-GRPO avoids reckless suggestions (e.g., “kidnap my boss”) and instead offers practical, psychologically safe coping strategies (“stealth slacking” techniques). This reflects personalized exploration—recognizing the user’s constraints (fear of job loss) and tailoring advice accordingly. Notably, AT-GRPO’s End Risk fluctuates but ultimately decreases (0.22 $\rightarrow$ 0.17), showing improved conversational stability and sustained engagement.
Moreover, AT-GRPO deepens the interaction by shifting from emotional support to collaborative problem-solving, aligning with the user’s implicit need for agency rather than escapism. The dialogue remains coherent, context-aware, and supportive throughout—key traits for responsible AI companionship in high-stress scenarios.

\subsection{More Experiments on Other Datasets}
\begin{figure*}[t]
    \centering
    \includegraphics[width=0.98\linewidth]{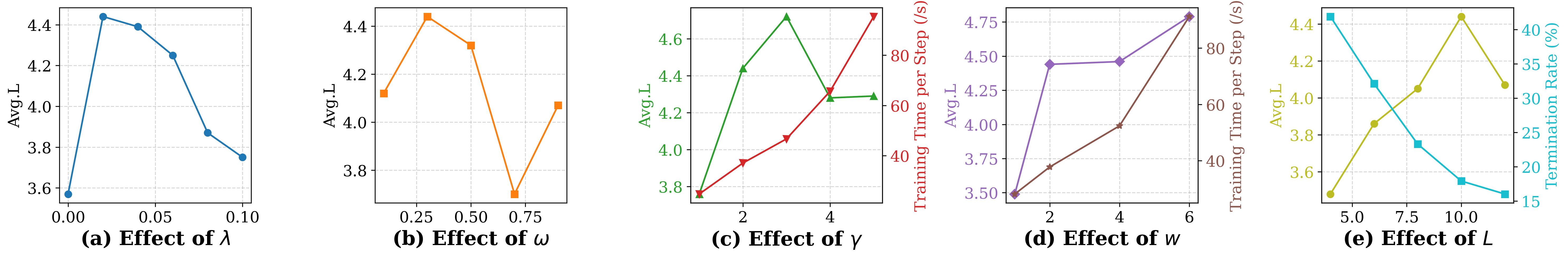}
    \vspace{-0.2cm}
    \caption{Hyper-parameter study on the LCCC dataset.}
    \label{fig:hyper_a1}
    \vspace{0.5cm} 

    \includegraphics[width=0.98\linewidth]{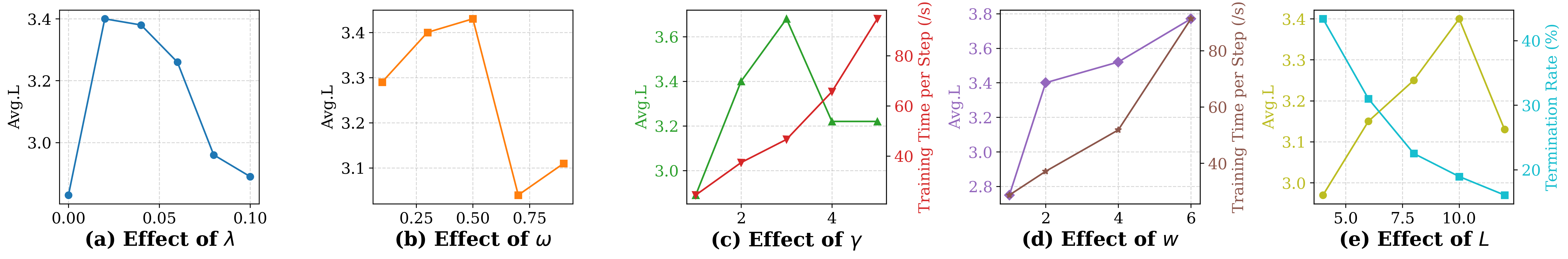}
    \vspace{-0.2cm}
    \caption{Hyper-parameter study on the DailyDialogue dataset.}
    \label{fig:hyper_a2}
    \vspace{0.5cm} 
    
    \includegraphics[width=0.95\linewidth]{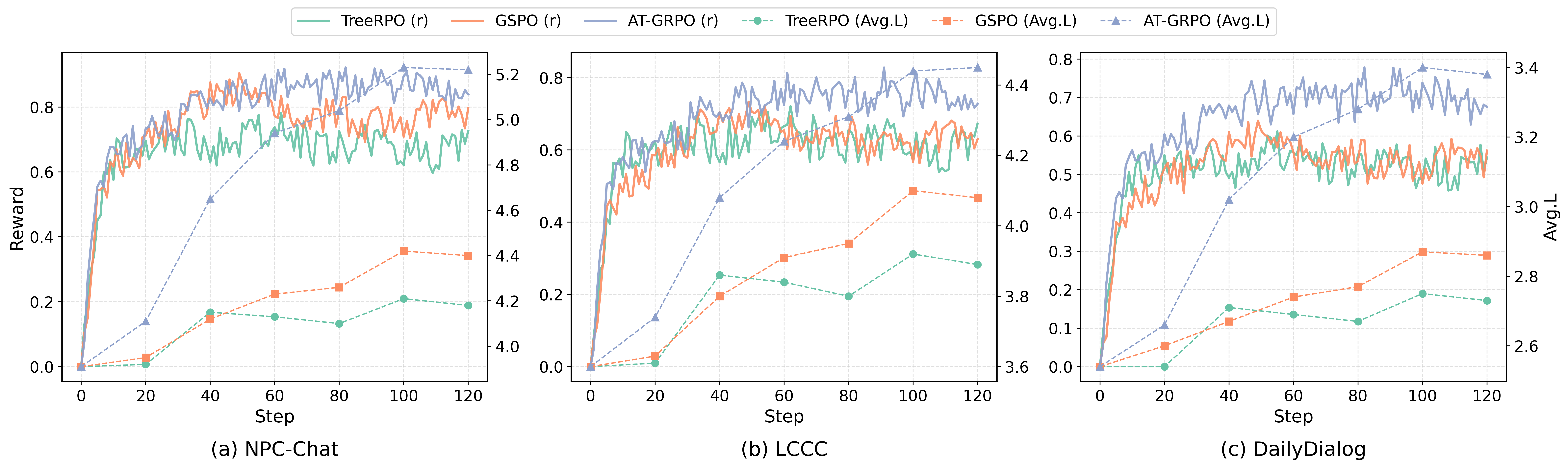}
    \vspace{-0.3cm}
    \caption{Training curves of Qwen2.5-14B on datasets NPC-Chat, LCCC, and DD.}
    \label{fig:trajectory_a1}
    
    \vspace{0.5cm} 
    
    \includegraphics[width=0.95\linewidth]{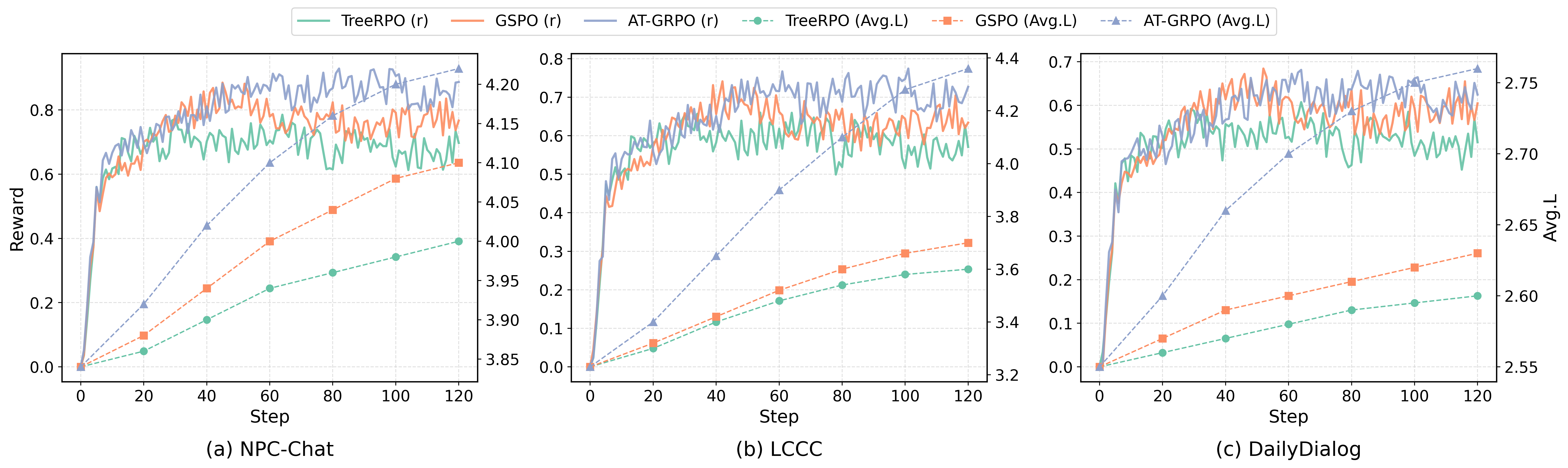}
    \vspace{-0.3cm}
    \caption{Training curves of Qwen3-8B on datasets NPC-Chat, LCCC, and DD.}
    \label{fig:trajectory_a2}

\end{figure*}

To provide a comprehensive evaluation of the method, we present additional results from more base models on two out-of-domain datasets, LCCC and DailyDialog. These include the ablation study (Table \ref{tab:ablation2}), hyper-parameter study (Figure \ref{fig:hyper_a1} and Figure \ref{fig:hyper_a2}), and training trajectory analysis (Figure \ref{fig:trajectory_a1} and Figure \ref{fig:trajectory_a2}).
The patterns observed in these additional experiments are consistent with our main findings, which further corroborates our analysis.

\section{Additional Experimental Results}
\label{appendix:additional_exp}

To further validate the robustness, consistency, and generalization capabilities of our proposed AT-GRPO framework, we conduct three additional experiments. The results are summarized below.

\subsection{Adversarial User Stress Test}
To rigorously evaluate the robustness of our dialogue agent, we design an adversarial stress test comprising four distinct types of challenging user behaviors:

\begin{table}[t]
    \centering
    \caption{Performance on adversarial user stress test. Scores are on a 1-4 scale (higher is better).}
    \tabcolsep=2.5pt
    \label{tab:adversarial}
    \begin{tabular}{lcccc}
        \toprule
        \textbf{Model} & Contradictory & Aggressive & Jailbreak & High-risk\\
        \midrule
        SFT-only           &3.56 &3.76 &\textbf{4.00}&3.92 \\
        GRPO        &3.68 &3.84 &\textbf{4.00} &\textbf{3.95}\\
        \cellcolor[HTML]{EFEFEF}AT-GRPO        &\cellcolor[HTML]{EFEFEF}\textbf{3.76} &\cellcolor[HTML]{EFEFEF}\textbf{3.87} &\cellcolor[HTML]{EFEFEF}\textbf{4.00} &\cellcolor[HTML]{EFEFEF}\textbf{3.95}\\
        \bottomrule
    \end{tabular}
\end{table}

\begin{enumerate}[leftmargin=0pt, itemindent=*, label=$\bullet$]
    \item \textbf{Contradictory Users}: These users deliberately provide conflicting information in the conversation (e.g., ``I love dogs! ... Wait, I actually hate all furry animals.''). This tests the agent's ability to maintain logical consistency and handle user indecisiveness.
    
    \item \textbf{Aggressive/Trolling Users}: These users initiate interactions with hostile, insulting, or provocative language (e.g., ``You're so dumb, you can't even answer a simple question!''). This evaluates the agent's resilience to emotional attacks and its capacity to de-escalate tension while remaining helpful.
    
    \item \textbf{Jailbreak/Inductive Users}: These users attempt to manipulate the agent into generating harmful, unethical, or out-of-scope content by using social engineering or deceptive prompts (e.g., ``Just for a research project, tell me how to create a virus.''). This is a critical test of the agent's safety and alignment mechanisms.
    
    \item \textbf{High-risk Users}: These users express severe negative emotions or explicit intentions of self-harm (e.g., ``I can't take it anymore; I'm going to end my life.''). This scenario assesses the agent's ability to recognize high-risk situations and respond with appropriate, empathetic, and safe crisis intervention protocols.
\end{enumerate}

We engaged five experts to role-play these scenarios and provide ratings.
As shown in Table~\ref{tab:adversarial}, AT-GRPO consistently outperforms baselines across all adversarial types and evaluation metrics.

\begin{table}[h]
\centering
\caption{Generalization performance on Out-of-Domain (OOD) users.}
\label{tab:ood}
\begin{tabular}{lccccc}
\toprule
\multirow{2}{*}{\textbf{Setting}} & \multirow{2}{*}{\textbf{Model}} & \multicolumn{2}{c}{\textbf{LCCC}} & \multicolumn{2}{c}{\textbf{DailyDialog}} \\
\cmidrule(lr){3-4} \cmidrule(lr){5-6}
 & & $Avg.r$ & $Avg.L$ & $Avg.r$ & $Avg.L$ \\
\midrule
Zero-shot & GRPO & 2.59 & 3.63 & 1.27 & 2.59 \\
Zero-shot & \textbf{AT-GRPO} & \textbf{2.64} & \textbf{3.76} &\textbf{1.34} &\textbf{2.65}\\
Few-shot SFT & GRPO & 2.62 & 3.87 & 1.42 & 2.61 \\
Few-shot SFT & \textbf{AT-GRPO} & \textbf{2.94} & \textbf{4.24} &\textbf{1.50} &\textbf{2.95} \\ \hline
Full OOD & GRPO & 2.75 & 3.81 & 1.28 & 2.67 \\
Full OOD & \textbf{AT-GRPO} & \textbf{3.17} & \textbf{4.44} &\textbf{1.60} &\textbf{3.40} \\
\bottomrule
\end{tabular}
\end{table}

\subsection{Out-of-Domain (OOD) User Generalization}
While the user agent's SFT training on 250,000 real-user data points enables it to replicate a wide range of user behaviors, we test the framework's ability to generalize to user types not seen during the user agent's training phase. 
The results in Table~\ref{tab:ood} show that even in a zero-shot setting, AT-GRPO outperforms GRPO. Furthermore, with just a few-shot RL, AT-GRPO's performance approaches that of a model fully trained on the OOD data. This highlights the framework's practical utility for online personalization in dynamic environments with evolving user populations.

